\documentclass[11pt]{article}

\usepackage[preprint]{acl}

\usepackage{times}
\usepackage{latexsym}
\usepackage[T1]{fontenc}
\usepackage[utf8]{inputenc}
\usepackage{microtype}
\usepackage{inconsolata}
\usepackage{graphicx}
\usepackage{booktabs}
\usepackage{makecell}
\usepackage{bm}
\usepackage{amssymb}
\usepackage{amsmath}
\usepackage{subcaption}
\usepackage{dblfloatfix}

\title{CHASE: Cache-Hole-Adapted Skip Exit for\\Looped State-Space Language Models}
\author{
  \begin{minipage}[t]{0.45\textwidth}
    \centering
    Zhenxuan Yu\thanks{
      E-mail: \texttt{24wr006m@rikkyo.ac.jp}
    } \\
    \textnormal{Rikkyo University} \\
    \vspace{0.5em}
    Yutaka Matsuo\thanks{
      Email: \texttt{\{t.kojima, matsuo, iwasawa\}@weblab.t.u-tokyo.ac.jp}
    } \\
    \textnormal{The University of Tokyo} \\
  \end{minipage}
  \hfill
  \begin{minipage}[t]{0.45\textwidth}
    \centering
    Takeshi Kojima\footnotemark[2] \\
    \textnormal{The University of Tokyo} \\
    \vspace{0.5em}
    Yusuke Iwasawa\footnotemark[2] \\
    \textnormal{The University of Tokyo} \\
  \end{minipage}
}

\begin{document}

\maketitle

\begin{abstract}
Recent work on looped language models suggests that many reasoning problems benefit from greater computational depth rather than from additional independent parameters. Existing studies, however, focus almost exclusively on Transformer backbones, leaving open whether this principle also applies to state-space language models. We investigate Looped Mamba and Looped Hybrid Mamba--Transformer architectures, which repeatedly apply a shared Mamba (or hybrid) block to introduce explicit finite-depth recurrent computation. On two controlled reasoning tasks---Mano (modular-arithmetic manipulation) and p-hop induction---Looped Mamba consistently outperforms parameter-matched non-looped baselines and, in several settings, matches or exceeds non-looped models of equal effective depth.
We then extend the study to language model pre-training under matched iso-parameter and iso-FLOPs protocols, which jointly disentangle the effects of parameter sharing and effective depth: looped models remain competitive on downstream benchmarks with substantially fewer distinct parameters, although deeper non-looped models retain an advantage in validation perplexity under strict iso-FLOPs comparisons.
Finally, we adapt Ouro's two-stage exit gate to Looped Mamba for threshold-controlled selection among recurrent-step outputs. Executing such exits on a state-space backbone, however, leaves the recurrent state without its deeper updates, and validation perplexity then degrades severely. We therefore introduce a cache-hole adaptation that aligns continued training with skipped-state inference. At the scales studied, the adapted model keeps perplexity close to full computation and matches or exceeds full-compute exit-state selection on downstream benchmarks while executing roughly half of the recurrent steps, which translates into measured inference speedups once the prefill is compute-bound.\footnote{Code will be released publicly.}
\end{abstract}


\section{Introduction}

Large Language Models (LLMs) have achieved remarkable progress in language understanding, text generation, and complex reasoning ~\citep{deepseekai2025deepseekv3technicalreport, yang2025qwen3technicalreport}. However, these advances have largely been driven by scaling model size, training data, and compute-intensive training and test-time inference ~\citep{hoffmann2022trainingcomputeoptimallargelanguage, snell2024scalingllmtesttimecompute}. This scaling paradigm introduces substantial challenges for practical deployment: model weights and intermediate activations increase memory consumption, while Transformer-based self-attention requires maintaining large key-value caches during long-context inference, leading to rapidly growing inference costs as context length increases ~\citep{beltagy2020longformerlongdocumenttransformer}. Improving model capability under fixed parameter, FLOP, or memory budgets has therefore become a central problem in the design of efficient LLM architectures. 

State Space Models (SSMs), particularly Mamba-2~\citep{gu2024mambalineartimesequencemodeling, dao2024transformers}, provide an efficient alternative to attention with favorable long-context scaling, and Hybrid Mamba-Transformer designs~\citep{lieber2024jambahybridtransformermambalanguage, waleffe2024empirical, blakeman2025nvidia} have further emerged as practical backbones for modern LLMs. Orthogonally, Looped Transformers reuse a shared block across depth to gain effective computational depth without extra parameters, and recent work shows that this benefits reasoning~\citep{dehghani2019universaltransformers, saunshi2025reasoning}.

Our core idea is to share the parameters of Mamba or hybrid blocks across layers and apply them repeatedly through explicit finite-depth loops, obtaining a deeper effective computational path without additional trainable parameters. We validate this design on controlled synthetic reasoning tasks (Mano and p-hop induction) that require compositional computation and recursive retrieval, and then on language-model pre-training under matched iso-parameter and iso-FLOPs protocols, which together separate the contributions of parameter sharing, additional compute, and effective depth. We further compare Looped Mamba with Looped Transformer and Looped Hybrid Mamba--Transformer on downstream reasoning benchmarks.

We then ask whether adaptive depth can also reduce inference cost. Adapting the post-training exit gate of Ouro~\citep{zhu2025scaling} to Looped Mamba already improves downstream accuracy over fixed-depth inference, but it saves no computation, because the gate only selects which recurrent step to read out. Actually skipping the deeper loops of an exited token exposes a failure mode specific to state-space backbones: the skipped steps never write the recurrent state, and the resulting \emph{cache holes} make perplexity diverge. We introduce CHASE (Cache-Hole--Adapted Skip Exit), which continues training on cache-hole-consistent forward passes so that the states seen during training match those induced by executable early exits.

The main contributions of this work are as follows.
First, to the best of our knowledge, we provide the first systematic study of explicit finite-depth looping for Mamba-based and Hybrid Mamba-Transformer language models, under controlled iso-parameter and iso-FLOPs comparisons against both Mamba and Transformer baselines.
Second, we present empirical results on controlled reasoning tasks (Mano, p-hop induction) and on language-model pre-training with downstream benchmarks, showing that recurrent depth improves compositional and recursive computation as well as language modeling efficiency.
Third, we identify the cache-hole mismatch that makes early-exit loop skipping fail on state-space backbones, and introduce CHASE, which removes it: skipping then stays within a few percent of the full-compute perplexity, matches or exceeds the full-compute exit gate on downstream benchmarks at about half the executed loops, and delivers measured prefill speedups.

\section{Related Work}
\paragraph{State Space Models and
  Mamba.}
State Space Models (SSMs), particularly Mamba and Mamba-2, offer a promising alternative to the efficiency bottlenecks of pure Transformer architectures~\citep{gu2024mambalineartimesequencemodeling}. Prior work has shown that Mamba-style models can match or outperform Transformers at small to medium scales in language modeling, while Mamba-2 further establishes a connection between SSMs and attention through structured state space duality and introduces a more efficient core layer design~\citep{dao2024transformers}. Owing to their favorable sequence-modeling and memory-efficiency characteristics compared with standard attention, Mamba-based architectures provide an attractive path toward efficient long-sequence language modeling and inference. Meanwhile, Hybrid Mamba-Transformer~\citep{lieber2024jambahybridtransformermambalanguage, waleffe2024empirical} architectures have begun to appear in recent LLM systems; for example, Nemotron 3 combines Mamba, Transformer, and Mixture-of-Experts components to improve throughput and support long-context modeling~\citep{blakeman2025nvidia}. These developments suggest that Mamba and hybrid Mamba-Transformer designs are becoming important components of modern efficient LLM architectures.

\paragraph{Recursive Mamba models.}
A closely related recent study, \citet{wang2026tinyrecursivereasoningmamba2}, replaces the attention blocks in the Tiny Recursive Model (TRM)~\citep{jolicoeurmartineau2025morerecursivereasoningtiny} with Mamba-2 attention-hybrid blocks and evaluates the resulting model on ARC-AGI, showing that Mamba-2 hybrid operators can preserve, and in some settings improve, recursive reasoning capability within a tiny task-specialized model.
\begin{figure*}[t]
  \centering
  \includegraphics[width=\textwidth]{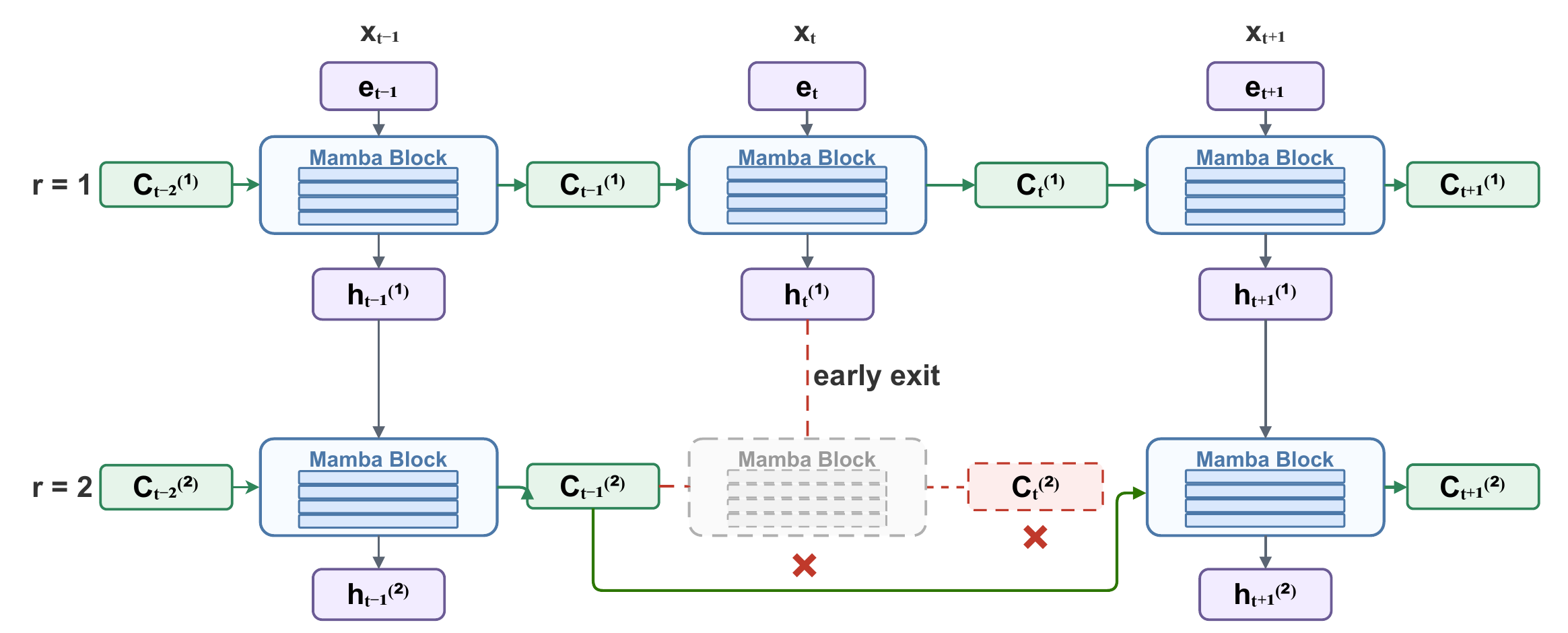}
  \caption{Illustration of adaptive exit and the resulting cache-hole problem in Looped Mamba. For clarity, each shared Mamba block is illustrated with four Mamba layers, while $C_t^{(r)}$ denotes the collection of layer-wise recurrent states after processing token $x_t$ at recurrent step $r$. In Ouro Stage-1 training, the exit gate is evaluated from the token hidden state $h_t^{(r)}$ after each non-final recurrent step. Under dense execution, all tokens are still processed through every recurrent step, so the corresponding Mamba recurrent-state updates are fully produced even when an earlier hidden state is selected for readout. When the learned exit decision is instead executed as an actual skip, a token that exits after step $r=1$ retains its already computed hidden state $h_t^{(1)}$, but is not processed by the deeper Mamba block. Consequently, $h_t^{(2)}$ is not computed and the updated recurrent state $C_t^{(2)}$ is never produced. The next active token is therefore processed from a deeper-loop cache that has not incorporated the skipped token, creating a \emph{cache hole} and a recurrent-state trajectory that differs from dense Stage-1 training. CHASE addresses this mismatch by continuing training with cache-hole-consistent forward passes, so that the hidden states and recurrent states encountered during training match those induced by executable early-exit inference.}
  \label{fig:chase_cache_hole}
\end{figure*}
\paragraph{Other Loop models.}
Looped or recurrent-depth architectures have a long history in sequence modeling. Universal Transformers apply the same transformation recurrently over sequence representations, combining the parallelism and global receptive field of Transformers with a recurrent inductive bias; they also introduce adaptive computation mechanisms for dynamic halting~\citep{dehghani2019universaltransformers}. A related line studies implicit SSMs that self-iterate toward a fixed point~\citep{schoene2025implicitlanguagemodelsrnns}, in contrast to the explicit finite-depth loops we study. More recently, \citet{saunshi2025reasoning} study Looped Transformers for reasoning and show that a $k$-layer Transformer looped $L$ times can nearly match, and sometimes outperform, a non-looped $kL$-layer model on synthetic and downstream reasoning tasks, while using far fewer distinct parameters. These results suggest that many reasoning tasks may require effective depth more than independent parameters. However, these studies focus primarily on Transformer backbones. In contrast, we investigate whether the same looped-depth principle applies to Mamba and Hybrid Mamba-Transformer architectures. Building on this line of work, Ouro~\citep{zhu2025scaling} further introduces an entropy-regularized exit gate that learns a token-level distribution over recurrent steps for adaptive exit-state selection on Looped Transformers; we adapt this mechanism to Looped Mamba and Hybrid Mamba-Transformer backbones and study its selected-depth--performance trade-off.

\paragraph{Looped models with subquadratic attention.}
LT2~\citep{deng2026lt2} replaces the quadratic softmax attention of Looped Transformers with subquadratic token mixers, comparing linear-attention, sparse-attention, and hybrid variants at the 0.6B--1.3B scale; Mamba-2 is one of the linear mixers it evaluates. LT2 keeps the number of loop iterations fixed during pre-training and leaves adaptive computation to future work, citing the non-differentiable halting rule, the sensitivity to the pondering weight, and the throughput loss caused by ragged halting. We instead study adaptive exit-state selection for looped state-space models: we learn a token-level exit gate for Looped Mamba and introduce a cache-hole adaptation (CHASE) that keeps the recurrent state consistent when deeper loops are skipped, so that early exits translate into actual compute savings. We further provide controlled iso-parameter and iso-FLOPs comparisons and study hybrid Mamba--Transformer backbones.

\section{Proposed Method}
\label{sec:method}
\subsection{Overall Architecture}

\paragraph{Looped Mamba and hybrid backbones.}
Let $M_{\theta}$ denote an $N$-layer Mamba-2 or Hybrid
Mamba--Transformer backbone. Given a length-$L$ token sequence $x$,
the dense fixed-loop forward is
\begin{equation}
\label{eq:looped_backbone}
\begin{split}
h^{(0)} &= \operatorname{emb}(x), \\
h^{(r)} &= M_{\theta}\bigl(h^{(r-1)}\bigr),
\quad r = 1,\ldots,R.
\end{split}
\end{equation}
We denote this model by $N\otimes R$, where $N$ is the number of
independently parameterized layers and $R$ is the number of recurrent
steps. Reusing the same backbone gives an effective depth of $NR$:
the parameter budget scales with $N$, whereas the backbone FLOPs scale
with $NR$. Thus, $N\otimes R$ is iso-parameter with $N\otimes1$ and
iso-FLOPs with the non-looped $NR\otimes1$ model.

For the hybrid variant, a small subset of Mamba-2 layers is replaced
by causal self-attention layers, and the entire mixed backbone is
shared and reapplied at every recurrent step. The exact attention
locations and architecture configurations are provided in
Appendix~\ref{app:mamba2_hybrid_details}.

Base pretraining applies next-token supervision only to the final
recurrent step. With
$p_{\theta}^{(r)}(\cdot\mid x_{\leq i})
=\operatorname{softmax}(\operatorname{lmhead}(h_i^{(r)}))$, where the same
language-modeling head is applied at every recurrent step, the
objective is
\begin{equation}
\begin{aligned}
\mathcal{L}_{\mathrm{final}}
&= -\frac{1}{L-1}\sum_{i=1}^{L-1}
\\[-0.15em]
&\quad \mathbb{E}_{x}\!\left[
\log p_{\theta}^{(R)}
\!\left(x_{i+1}\mid x_{\leq i}\right)
\right].
\end{aligned}
\label{eq:final_loop_pretraining}
\end{equation}

\subsection{Adaptive Exit and Cache-Hole Adaptation}
\label{sec:cache_hole_adaptation}

\paragraph{Stage-1 adaptive exit training.}
Following Ouro~\citep{zhu2025scaling}, we attach a lightweight
token-level exit gate to the recurrent backbone. Let
$x=(x_1,\ldots,x_L)$ denote a training sequence containing $L$
non-padding tokens. At recurrent step $r$, the language model defines
the token-level next-token loss for $i=1,\ldots,L-1$ as
\begin{equation}
\ell_i^{(r)}
= -\log p_{\theta}^{(r)}
\!\left(x_{i+1}\mid x_{\leq i}\right),
\label{eq:per_step_token_loss}
\end{equation}
where $\theta$ denotes the embedding, recurrent backbone, and
language-modeling head parameters.

For every non-final recurrent step $r=1,\ldots,R-1$, an affine exit gate with
parameters $\phi=(\bm{w},b)$ predicts the instantaneous exit
probability
\begin{equation}
\lambda_i^{(r)}
= \sigma\!\left(\bm{w}^{\top}h_i^{(r)}+b\right),
\label{eq:exit_gate}
\end{equation}
where $h_i^{(r)}$ is the hidden state of token $i$ after recurrent
step $r$. These instantaneous probabilities induce a distribution
over the token's exit step:
\begin{equation}
\pi_i^{(r)}
=
\begin{cases}
\lambda_i^{(r)}
\displaystyle
\prod_{j=1}^{r-1}
\left(1-\lambda_i^{(j)}\right),
& r<R,\\[3mm]
\displaystyle
\prod_{j=1}^{R-1}
\left(1-\lambda_i^{(j)}\right),
& r=R.
\end{cases}
\label{eq:exit_distribution}
\end{equation}
The final recurrent step therefore receives all remaining probability
mass, and $\sum_{r=1}^{R}\pi_i^{(r)}=1$.

Let
\begin{equation}
H(\bm{\pi}_i)
=
-
\sum_{r=1}^{R}
\pi_i^{(r)}
\log\!\left(\pi_i^{(r)}+\epsilon\right)
\label{eq:exit_entropy}
\end{equation}
denote the entropy of the exit distribution for token $i$. The
original Stage-1 objective is
\begin{equation}
\begin{aligned}
\mathcal{L}_{\mathrm{S1}}(x)
&= \frac{1}{L-1}\sum_{i=1}^{L-1}
\sum_{r=1}^{R}\pi_i^{(r)}\ell_i^{(r)}
\\[-0.15em]
&\quad -\frac{\beta}{L-1}\sum_{i=1}^{L-1}
H(\bm{\pi}_i),
\end{aligned}
\label{eq:stage1}
\end{equation}
where $\beta$ controls the strength of entropy regularization. The
expected token loss encourages the gate to assign probability to
recurrent steps that provide accurate predictions, while the entropy
term prevents the exit distribution from prematurely collapsing to a
single depth.

Here $\theta$ warm-starts from a fixed-loop pretrained checkpoint and is
optimized jointly with the newly attached gate $\phi$; initialization and
optimizer details are given in Appendix~\ref{app:exitgate_training}.

Figure~\ref{fig:chase_cache_hole} gives a concrete cache-level view of this
process. For clarity, each shared Mamba block is drawn as four stacked Mamba
layers and reused across recurrent steps. At the first recurrent step, the
gate reads the token hidden state $h_t^{(1)}$ and may select it as the exit
state. Dense Stage-1 training nevertheless continues to execute the deeper
shared block, so every layer-wise recurrent state is updated. If that decision
is instead executed as a real skip, token $x_t$ bypasses the second recurrent
step and the corresponding state collection $C_t^{(2)}$ is never written.
The next token therefore encounters a deeper-loop cache that omits the
contribution of $x_t$, producing the cache-hole mismatch addressed by CHASE.

\paragraph{Motivation: the cache-hole mismatch.}
We distinguish \emph{dense execution}, which runs every token through all
$R$ recurrent steps and uses the gate only to select the output step, from
\emph{state-skipping execution} (\emph{skip}), in which a token that exits
after step $r$ is dropped from all subsequent scans: its hidden state is
frozen and it no longer contributes to deeper convolutional or SSM state
updates. Dense Stage-1 training therefore only ever exposes the model to
complete state trajectories, while skip inference leaves the deeper updates
of exited tokens missing, or \emph{cache holes}.

\paragraph{Cache-hole-consistent forward.}
We address this mismatch by continuing training from the original
Stage-1 checkpoint, including its learned exit-gate parameters. Each
training minibatch follows either the original dense Stage-1 forward
or a cache-hole-consistent forward. Let $\rho$ denote the probability
of selecting the cache-hole branch, and let
$q\sim\mathcal{D}_{\mathrm{hole}}$ be one exit threshold sampled for
the corresponding minibatch.

For the cache-hole branch, let
$\mathcal{A}_r\subseteq\{1,\ldots,L\}$ denote the set of tokens that
remain active when entering recurrent step $r$, with
$\mathcal{A}_1=\{1,\ldots,L\}$. Only active tokens are packed in their
original causal order and processed by the recurrent backbone:
\begin{equation}
\begin{aligned}
\widetilde{\bm{u}}^{(r)}
&= M_{\theta}\!\left(
\operatorname{Pack}_{\mathcal{A}_r}
\!\left(\widetilde{\bm{h}}^{(r-1)}\right)
\right),
\\[-0.15em]
\widetilde{\bm{h}}^{(r)}
&= \operatorname{Scatter}_{\mathcal{A}_r}\!\left(
\widetilde{\bm{u}}^{(r)},
\widetilde{\bm{h}}^{(r-1)}
\right),
\end{aligned}
\label{eq:hole_state_update}
\end{equation}
where
$\widetilde{\bm{h}}^{(0)}=\operatorname{emb}(x)$ and
$\widetilde{\bm{u}}^{(r)}$ denotes the packed survivor updates.
The packing operation removes previously exited tokens while
preserving the relative causal order of the surviving tokens. The
scatter operation writes the updated survivor states back to their
original token positions and leaves the hidden states of exited tokens
unchanged. Thus, an exited token is absent from all deeper
convolutional windows and SSM scans, and its hidden state remains
frozen after exit.

After recurrent step $r<R$, the gate is evaluated on the resulting
cache-hole-consistent hidden state:
\begin{equation}
\widetilde{\lambda}_i^{(r)}
=
\operatorname{sg}
\left[
\sigma\!\left(
\bm{w}^{\top}
\widetilde{h}_i^{(r)}
+b
\right)
\right],
\label{eq:hole_gate}
\end{equation}
where $\operatorname{sg}(\cdot)$ denotes stop-gradient. Substituting
$\widetilde{\lambda}_i^{(r)}$ into
Eq.~\eqref{eq:exit_distribution} gives the exit distribution
$\widetilde{\pi}_i^{(r)}$ along the cache-hole-consistent trajectory.
Its cumulative exit probability is
\begin{equation}
\widetilde{F}_i^{(r)}
=
\sum_{j=1}^{r}
\widetilde{\pi}_i^{(j)}.
\label{eq:hole_cdf}
\end{equation}
The realized exit step under threshold $q$ is
\begin{equation}
\begin{aligned}
\widetilde{\mathcal{E}}_i(q)
&= \{\,r\in[R]:\widetilde{F}_i^{(r)}\geq q\,\},
\\[-0.15em]
\widetilde{z}_i(q)
&= \min\!\left(\widetilde{\mathcal{E}}_i(q)\cup\{R\}\right),
\end{aligned}
\label{eq:hole_exit}
\end{equation}
so that any token still active at the final recurrent step exits at $R$.
Tokens that exit at step $r$ are removed from $\mathcal{A}_{r+1}$.

\paragraph{Exit-depth supervision on holed states.}
The realized exit decision is a hard, one-hot exit distribution with zero
entropy, so the Stage-1 objective of Eq.~\eqref{eq:stage1} applied to it
reduces to the cross-entropy at the token's realized exit depth:
\begin{equation}
\mathcal{L}_{\mathrm{hole}}(x;q)
=
\frac{1}{L-1}
\sum_{i=1}^{L-1}
\widetilde{\ell}_i^{\left(\widetilde{z}_i(q)\right)},
\label{eq:hole_loss}
\end{equation}
where $\widetilde{\ell}_i^{(r)}$ is computed from the hidden state obtained
under the cache-hole-consistent trajectory. Each token is thus supervised
only at its realized exit depth, rather than by balanced deep supervision
over all steps.

In expectation over the stochastic branch selection, our
cache-hole-adaptation objective is
\begin{equation}
\begin{aligned}
\mathcal{L}_{\mathrm{CHASE}}
&= (1-\rho)\,
\mathbb{E}_{x}\!\left[\mathcal{L}_{\mathrm{S1}}(x)\right]
\\[-0.15em]
&\quad +\rho\,
\mathbb{E}_{x,\,q}
\!\left[\mathcal{L}_{\mathrm{hole}}(x;q)\right].
\end{aligned}
\end{equation}
Dense minibatches update both $\theta$ and $\phi$ through
Eq.~\eqref{eq:stage1}; cache-hole minibatches apply stop-gradient to the hard
exit decisions, so Eq.~\eqref{eq:hole_loss} updates the backbone and the
language-modeling head but not the gate, which keeps learning from the dense
branch. Inference reuses the cumulative-threshold and survivor-packing
semantics of the cache-hole branch, so continued training and executable
early exits induce the same state trajectories.

The full training sequence is fixed-loop pretraining, entropy-regularized
Stage-1 exit-gate training, and the proposed cache-hole adaptation; the
optional gate-only calibration stage that follows is unchanged from Ouro
(Appendix~\ref{app:exitgate_training}). We refer to the resulting method---the
cache-hole--adapted gate together with its executable skip inference---as
\textbf{CHASE} (Cache-Hole--Adapted Skip Exit).

\section{Looped Models on Simple Reasoning Tasks}
We evaluate Looped Mamba on two synthetic reasoning tasks, using the $N\otimes R$ notation of Section~\ref{sec:method} throughout, where $N$ is the number of independently parameterized Mamba layers and $R$ the number of recurrent steps.
\subsection{Mano Task}
\paragraph{Setup.}
We first evaluate knowledge manipulation on the Mano task~\citep{allenzhu2025physicslanguagemodels41}, a synthetic modular-arithmetic reasoning task in which the model must directly output the final answer, applying arithmetic rules modulo 23 and parsing the tree structure encoded by a prefix expression to compose multiple computations, without producing any explicit intermediate steps.
Difficulty is set by the maximum expression length $L$, i.e., the number of operations in each sample: expression lengths $\ell$ are sampled from $[1,L]$ during training, and evaluation uses only the hardest expressions with $\ell=L$.
We use three difficulty levels $L\in\{10,16,24\}$ and, for each, compare an iso-parameter non-looped baseline ($4\otimes1$, $6\otimes1$, $12\otimes1$) with its looped counterpart ($4\otimes6$, $6\otimes4$, $12\otimes2$); all three looped models have effective depth 24 and are therefore iso-FLOPs with a $24\otimes1$ non-looped reference, while using only $1/6$, $1/4$, and $1/2$ of its parameterized layers, respectively.
More details are provided in Appendix~\ref{app:mano}.

\paragraph{Results.}
Table~\ref{tab:mano_mamba} reports the exact-match accuracy on the Mano task. 
Under the iso-parameter comparison, Looped Mamba consistently outperforms its non-looped counterpart across all model sizes and difficulty levels. 
For example, on the hardest setting $L=24$, the $4\otimes1$, $6\otimes1$, and $12\otimes1$ baselines achieve only 7.50\%, 8.00\%, and 22.70\% accuracy, respectively, whereas their looped counterparts improve to 99.35\%, 59.85\%, and 38.40\%. 
These results indicate that recurrently reusing Mamba layers substantially improves the model's ability to parse arithmetic tree structures and compose modular computations, without increasing the number of trainable parameters.

Under the same effective depth of 24, i.e., the strict iso-FLOPs setting, the best looped model, $4\otimes6$, outperforms the $24\otimes1$ non-looped baseline on all difficulty levels, while using only one sixth of the parameterized layers.
In particular, on $L=24$, $4\otimes6$ improves accuracy from 37.45\% to 99.35\%. 
This comparison shows that the gain is not simply due to using more computation: with matched effective depth, parameter sharing through recurrent loops can provide a stronger inductive bias for knowledge manipulation than increasing the number of unshared layers. 
Moreover, among the effective-depth-24 looped models, performance improves as the number of recurrent steps increases, from $12\otimes2$ to $6\otimes4$ and further to $4\otimes6$. 
\begin{table}[t]
\centering
\resizebox{\columnwidth}{!}{%
\begin{tabular}{lcccc}
\toprule
\textbf{Model} 
& \makecell{\textbf{Params /}\\\textbf{FLOPs}} 
& \textbf{$L=10$} 
& \textbf{$L=16$} 
& \textbf{$L=24$} \\
\midrule
\multicolumn{5}{l}{\textbf{Baseline model}} \\
Base $(24 \otimes 1)$ 
& $24\times\,/\,24\times$ 
& 96.10 & 89.45 & 37.45 \\
\midrule
\multicolumn{5}{l}{\textbf{4-layer model}} \\
Base $(4 \otimes 1)$  
& $4\times\,/\,4\times$   
& 55.60 & 12.85 & 7.50 \\
Loop $(4 \otimes 6)$  
& $4\times\,/\,24\times$  
& \textbf{99.21} & \textbf{99.61} & \textbf{99.35} \\
\midrule
\multicolumn{5}{l}{\textbf{6-layer model}} \\
Base $(6 \otimes 1)$  
& $6\times\,/\,6\times$   
& 86.33 & 18.35 & 8.00 \\
Loop $(6 \otimes 4)$
& $6\times\,/\,24\times$
& \textbf{97.90} & \textbf{82.42} & \textbf{59.85} \\
\midrule
\multicolumn{5}{l}{\textbf{12-layer model}} \\
Base $(12 \otimes 1)$ 
& $12\times\,/\,12\times$ 
& 92.97 & 59.70 & 22.70 \\
Loop $(12 \otimes 2)$
& $12\times\,/\,24\times$
& \textbf{95.70} & \textbf{82.03} & \textbf{38.40} \\
\bottomrule
\end{tabular}%
}
\caption{Accuracy (\%) of looped/non-looped Mamba models on the Mano task. Bold indicates the better result within each iso-parameter group.}
\label{tab:mano_mamba}
\end{table}
\subsection{p-hop Task}
\label{sec:phop}
\paragraph{Setup.}We further evaluate the recursive retrieval capability of Looped Mamba on the p-hop induction task. This task is a multi-step generalization of the induction-head problem. Given a sequence $v=(v_1,\ldots,v_n)$, 1-hop induction starts from the last token $v_n$, finds its previous occurrence in the sequence, and outputs the token immediately following that occurrence. The p-hop task recursively applies this operation $p$ times, requiring the model to repeatedly backtrack through the context and retrieve the final answer.
We replace the Transformer backbone in the original p-hop experiments with a Mamba-2 backbone while keeping the same looped-model protocol. We use an alphabet of size 4, sequence length $n=256$, and evaluate two difficulty levels, $p=16$ and $p=32$. Following the Looped Transformer setup, we compare non-looped baselines, iso-parameter looped models, and iso-FLOPs baselines. Since the answer is one of four alphabet tokens, random guessing obtains roughly 25\% accuracy. Full dataset construction, training, and evaluation details are provided in Appendix~\ref{app:phop}.

\begin{table}[t]
\centering
\scriptsize
\setlength{\tabcolsep}{4pt}
\renewcommand{\arraystretch}{1.0}
\begin{tabular}{lccc}
\toprule
\textbf{Model} & \makecell{\textbf{Params /}\\\textbf{FLOPs}} & \makecell{$\boldsymbol{p{=}16,}$\\$\boldsymbol{n{=}256}$} & \makecell{$\boldsymbol{p{=}32,}$\\$\boldsymbol{n{=}256}$} \\
\midrule
Base $(12\otimes1)$ & $12\times/12\times$ & 99.95 & 99.87 \\
\midrule
\multicolumn{4}{l}{\textbf{1-layer model}} \\
Base $(1\otimes1)$  & $1\times/1\times$   & 38.32 & 41.08 \\
Loop $(1\otimes6)$  & $1\times/6\times$   & 99.51 & 98.78 \\
Loop $(1\otimes12)$ & $1\times/12\times$  & 99.70 & 99.89 \\
\midrule
\multicolumn{4}{l}{\textbf{2-layer model}} \\
Base $(2\otimes1)$  & $2\times/2\times$   & 63.51 & 67.22 \\
Loop $(2\otimes3)$  & $2\times/6\times$   & 99.55 & 98.39 \\
Loop $(2\otimes6)$  & $2\times/12\times$  & 99.93 & 99.84 \\
\midrule
\multicolumn{4}{l}{\textbf{3-layer model}} \\
Base $(3\otimes1)$  & $3\times/3\times$   & 87.18 & 77.38 \\
Loop $(3\otimes2)$  & $3\times/6\times$   & 99.74 & 98.55 \\
Loop $(3\otimes4)$  & $3\times/12\times$  & 99.92 & 99.80 \\
\midrule
\multicolumn{4}{l}{\textbf{4-layer model}} \\
Base $(4\otimes1)$  & $4\times/4\times$   & 97.99 & 89.39 \\
Loop $(4\otimes3)$  & $4\times/12\times$  & 99.89 & 99.84 \\
\midrule
\multicolumn{4}{l}{\textbf{6-layer model}} \\
Base $(6\otimes1)$  & $6\times/6\times$   & 99.81 & 99.38 \\
Loop $(6\otimes2)$  & $6\times/12\times$  & \textbf{99.97} & \textbf{99.91} \\
\bottomrule
\end{tabular}
\caption{Accuracy (\%) of looped and non-looped Mamba-2 models on p-hop induction. Params/FLOPs gives the relative number of parameterized layers and the effective depth, respectively. Bold marks the best result in each column.}
\label{tab:phop_mamba}
\end{table}

\paragraph{Results.}Table~\ref{tab:phop_mamba} reports the results of Looped Mamba on p-hop induction. Shallow non-looped Mamba models struggle on this task: the $1 \otimes 1$ model obtains only 38.32\% and 41.08\% accuracy for $p=16$ and $p=32$, respectively, while the $2 \otimes 1$ model reaches 63.51\% and 67.22\%. Recurrent parameter sharing closes this gap rapidly: $1 \otimes 6$ already reaches 99.51\% and 98.78\%, and $1 \otimes 12$ further improves to 99.70\% and 99.89\%. At matched effective depth 12, $2 \otimes 6$, $3 \otimes 4$, $4 \otimes 3$, and $6 \otimes 2$ all achieve at least 99.80\% on both difficulty levels and are competitive with the much deeper $12 \otimes 1$ baseline.

These results suggest that p-hop induction primarily requires recursive computational depth rather than a large number of distinct parameters. Even one- or two-layer Mamba-2 backbones, when given sufficient effective depth via looping, can learn recursive backtracking and answer retrieval. Consistent with the findings for Looped Transformers, Looped Mamba substantially outperforms the iso-parameter baseline and matches or exceeds the performance of deeper non-looped baselines. This indicates that Mamba's selective state-update mechanism can also benefit from recurrent parameter sharing and can perform multi-step latent retrieval with far fewer parameters.

\section{Language Modeling with Looped Models}
\subsection{Setup}
\paragraph{Models}
We pre-trained looped language models based on LLaMA-2 Transformer \cite{touvron2023llama2openfoundation} and Mamba-2 architectures, with model sizes ranging from 50M to 370M parameters. All models follow the GPT-like autoregressive decoder pre-training methodology \cite{NEURIPS2020_1457c0d6}. Table~\ref{tab:isoflops_lm} compares LLaMA-2 Transformer and Mamba-2 models under iso-parameter and iso-FLOPs settings, while Table~\ref{tab:mamba2_exitgate_full} evaluates LLaMA-2, Mamba-2, Mamba-2 Hybrid, and exit-gate variants.
\paragraph{Training settings}
Models in Table~\ref{tab:isoflops_lm} are pre-trained for 100B tokens, corresponding to 50,000 iterations, from the Nemotron-ClimbMix dataset \cite{diao2025climb}. For the models in Table~\ref{tab:mamba2_exitgate_full}, we use a four-stage training procedure. First, all models are pre-trained for 150B tokens, corresponding to 75,000 iterations, from the same dataset. Second, the exit-gate variants are continued with Exit-Gate Stage~1 training for 5.2B tokens, corresponding to 2,500 iterations. Third, we apply cache-hole adaptation (CHASE) from the Stage-1 weights: we continue training with holed forward passes that hole a fraction $\mathrm{hp}{=}0.5$ of positions and, for each hole, sample a target exit depth uniformly in $[1.85,3.45]$ over the $R{=}4$ recurrent steps (depth-space sampling); the remaining hyperparameters and per-scale step counts are given in Appendix~\ref{app:chase_training}. Fourth, Exit-Gate Stage~2 further trains the gate for 1.2B tokens, corresponding to 600 iterations. The \emph{loop4-Ouro} baseline omits the cache-hole adaptation step (the third stage) and calibrates Stage~2 directly on the Stage-1 weights; \emph{loop4-Ouro} and \emph{loop4-CHASE} therefore share the same Stage-1 backbone and the same Stage-2 gate training and differ only in whether cache-hole adaptation is applied, which isolates its effect. We use the Mistral-7B-v0.1 tokenizer \cite{jiang2023mistral7b} with a vocabulary size of 32,000 and set the context length to 2048 tokens. All experiments are conducted on 8 H100 GPUs with an effective global batch size of 1024 sequences. Architecture-specific configurations, optimizer settings, and detailed results are provided in Appendix~\ref{app:training_details}.
\paragraph{Evaluation}
For downstream evaluation, we report performance on a set of commonsense and reasoning benchmarks, including PIQA~\citep{bisk2020piqa}, BoolQ~\citep{clark2019boolq}, WinoGrande~\citep{sakaguchi2021winogrande}, OpenBookQA~\citep{OpenBookQA2018}, HellaSwag~\citep{zellers2019hellaswag}, ARC-Challenge, and ARC-Easy~\citep{clark2018think}. We follow a fixed few-shot evaluation protocol: PIQA, BoolQ, WinoGrande, and OpenBookQA are evaluated with 5-shot prompting; HellaSwag with 10-shot prompting; ARC-Challenge with 25-shot prompting; and ARC-Easy with 8-shot prompting.
\subsection{iso-FLOPs Evaluation}
Table~\ref{tab:isoflops_lm} reports the complete validation-perplexity and per-task downstream results. 
The table supports two complementary comparisons. 
First, under the iso-parameter setting, looped models consistently improve validation perplexity over their non-looped counterparts with the same number of parameterized layers, for both Mamba-2 and LLaMA-2 backbones. 
This indicates that recurrently reusing the same backbone improves language modeling performance when the trainable parameter budget is fixed. For the LLaMA-2 Transformer backbone, each parameterized layer contains both causal self-attention and a gated MLP, whereas each Mamba-2 layer uses a Mamba-2 mixer without an external MLP. 
In Table~\ref{tab:isoflops_lm}, both backbones are compared using the same layer counts $N\in\{6,12,24\}$ and the same loop notation $N\otimes R$. 
Thus, the comparison aligns the number of parameterized layers and effective depth, while architectural details such as attention, MLPs, and Mamba mixing remain backbone-specific.

Second, under the strict iso-FLOPs setting, the deeper non-looped $24\otimes1$ models still achieve the best validation perplexity. Although looped models match the effective depth of $24\otimes1$, they use substantially fewer distinct parameters and do not fully close the perplexity gap.
This behavior is consistent with prior observations for Looped Transformers, where looped language models improve over iso-parameter baselines but remain worse than iso-FLOPs non-looped models in perplexity~\citep{saunshi2025reasoning}.

Notably, downstream performance does not follow validation perplexity exactly. 
Despite their worse perplexity relative to the $24\otimes1$ baselines, looped models remain competitive on the downstream average and in some cases outperform the deeper non-looped baseline. 
This suggests that recurrent depth may provide a parameter-efficient inductive bias that is not fully captured by perplexity alone, although the strength of this effect depends on the backbone and task.
\begin{table*}[t]
\centering
\footnotesize
\setlength{\tabcolsep}{4pt}
\renewcommand{\arraystretch}{0.95}
\begin{tabular}{llcccccccccc}
\toprule
\textbf{Arch.}
& \textbf{Model}
& \textbf{Params / FLOPs}
& \textbf{PPL}
& \textbf{PIQA}
& \textbf{BoolQ}
& \textbf{WG}
& \textbf{OBQA}
& \textbf{HS}
& \textbf{ARC-C}
& \textbf{ARC-E}
& \textbf{Avg.} \\
\midrule
\multicolumn{12}{l}{\textbf{Mamba-2}} \\
Mamba-2 & Base $(6\otimes1)$  & $6\times / 6\times$   & 21.96 & 51.52 & 39.57 & 50.28 & 26.80 & 26.36 & 27.39 & 25.25 & 35.31 \\
Mamba-2 & Loop $(6\otimes4)$  & $6\times / 24\times$  & 19.98 & 52.77 & 50.61 & 51.14 & 28.00 & 25.91 & 27.99 & 25.29 & 37.39 \\
Mamba-2 & Base $(12\otimes1)$ & $12\times / 12\times$ & 18.66 & 52.88 & 61.93 & 48.86 & 24.20 & 26.02 & 27.39 & 25.59 & 38.12 \\
Mamba-2 & Loop $(12\otimes2)$ & $12\times / 24\times$ & 18.21 & 49.89 & 62.11 & 49.49 & 23.40 & 25.80 & \textbf{28.58} & \textbf{26.85} & 38.02 \\
Mamba-2 & Base $(24\otimes1)$ & $24\times / 24\times$ & \textbf{16.15} & 51.90 & 57.37 & 48.07 & 26.40 & 25.46 & 28.41 & 25.38 & 37.57 \\
\midrule
\multicolumn{12}{l}{\textbf{LLaMA-2}} \\
Transformer & Base $(6\otimes1)$  & $6\times / 6\times$   & 21.48 & 52.94 & 59.30 & 47.99 & 25.80 & 25.75 & 28.92 & 26.64 & 38.19 \\
Transformer & Loop $(6\otimes4)$  & $6\times / 24\times$  & 19.75 & 52.67 & 57.98 & 49.64 & 27.80 & \textbf{26.17} & 28.24 & 26.64 & 38.45 \\
Transformer & Base $(12\otimes1)$ & $12\times / 12\times$ & 18.53 & 51.47 & 62.05 & 50.59 & 26.00 & 26.02 & 29.01 & 25.84 & 38.71 \\
Transformer & Loop $(12\otimes2)$ & $12\times / 24\times$ & 18.06 & \textbf{52.72} & \textbf{62.17} & 50.59 & 28.60 & 25.62 & 27.05 & 26.56 & \textbf{39.04} \\
Transformer & Base $(24\otimes1)$ & $24\times / 24\times$ & 16.40 & 50.92 & 49.54 & \textbf{51.14} & \textbf{28.80} & 26.02 & 28.75 & 25.29 & 37.21 \\
\bottomrule
\end{tabular}
\caption{Complete iso-FLOPs evaluation of Mamba-2 and LLaMA-2 Transformer models trained on 100B tokens. PPL denotes validation perplexity, and Avg. denotes the average score over the seven downstream tasks shown in the table.}
\label{tab:isoflops_lm}
\end{table*}
\subsection{iso-Parameter Evaluation and Adaptive Exit}
\label{sec:iso_parameter_exit}

\begin{figure}[t]
\centering
\includegraphics[width=\columnwidth]{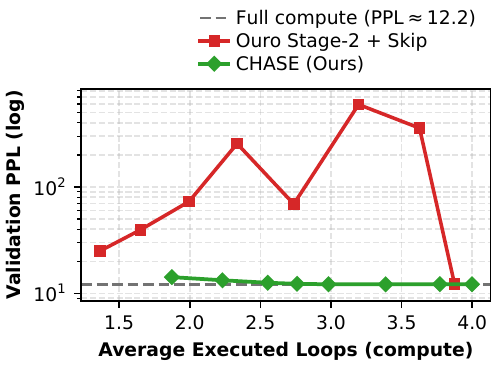}
\caption{Validation perplexity under gate+skip inference against the average number of executed loops (370M; log scale). As compute is reduced, the stock Ouro Stage-2 gate's perplexity explodes (by up to $+4788\%$; the Missing-State / cache-hole effect), whereas CHASE stays within $+3$--$4\%$ of the full-compute perplexity (grey dashed).}
\label{fig:ppl_skip_compute}
\end{figure}

\begin{table*}[t]
\centering
\resizebox{\textwidth}{!}{
\begin{tabular}{lcccccccccccc}
\toprule
\textbf{Model} & \textbf{Exit} & \textbf{$q$} & \textbf{Exec.} & \textbf{PPL} & \textbf{PIQA} & \textbf{BoolQ} & \textbf{WG} & \textbf{OBQA} & \textbf{HS} & \textbf{ARC-C} & \textbf{ARC-E} & \textbf{Avg.} \\
\midrule
\multicolumn{13}{l}{\textbf{140M LLaMA-2}} \\
baseline & full & -- & 1.00 & 15.96 & 50.98 & 56.51 & 50.59 & \textbf{27.60} & 25.48 & \textbf{29.44} & 23.91 & 37.79 \\
loop4 & full & -- & 4.00 & \textbf{14.75} & \textbf{51.09} & \textbf{60.83} & \textbf{51.70} & 25.60 & \textbf{25.88} & 27.56 & \textbf{26.18} & \textbf{38.41} \\
\midrule
\multicolumn{13}{l}{\textbf{140M Mamba-2}} \\
baseline & full & -- & 1.00 & 15.80 & 49.95 & 55.38 & \textbf{51.22} & \textbf{28.00} & 26.11 & \textbf{29.44} & 25.55 & 37.95 \\
loop2 & full & -- & 2.00 & 15.41 & 51.52 & 39.66 & 49.01 & \textbf{28.00} & 26.06 & 27.82 & \textbf{26.94} & 35.57 \\
loop4 & full & -- & 4.00 & \textbf{14.68} & 52.34 & 58.87 & 48.22 & 27.00 & 25.21 & 28.07 & 24.83 & 37.79 \\
loop4-Ouro & full & 0.8 & 4.00 & 15.54 & \textbf{52.99} & \textbf{62.23} & 50.20 & 26.00 & 26.01 & 28.92 & 25.72 & \textbf{38.86} \\
loop4-CHASE (Ours) & skip & 0.2 & 2.15 & 16.45 & 52.61 & 62.17 & 50.99 & 25.80 & \textbf{26.18} & 28.75 & 25.21 & 38.82 \\
\midrule
\multicolumn{13}{l}{\textbf{140M Mamba-2 Hybrid}} \\
baseline & full & -- & 1.00 & 15.27 & 51.09 & 37.83 & \textbf{50.28} & \textbf{27.00} & \textbf{25.71} & 28.50 & 26.35 & 35.25 \\
loop4 & full & -- & 4.00 & \textbf{14.37} & \textbf{51.31} & 45.87 & 47.43 & 26.80 & 25.07 & \textbf{28.92} & 26.09 & 35.93 \\
loop4-Ouro & full & 0.8 & 4.00 & 15.07 & 50.38 & \textbf{54.80} & 49.25 & 26.40 & 25.39 & 27.73 & \textbf{26.68} & \textbf{37.23} \\
\midrule
\multicolumn{13}{l}{\textbf{370M Mamba-2}} \\
baseline & full & -- & 1.00 & 12.72 & 48.20 & 61.30 & 49.60 & 27.20 & 24.60 & 28.90 & 25.70 & 37.93 \\
loop4 & full & -- & 4.00 & \textbf{12.06} & 51.90 & \textbf{61.96} & 48.15 & \textbf{27.80} & 24.72 & \textbf{29.10} & 25.93 & \textbf{38.51} \\
loop4-Ouro & full & 0.8 & 4.00 & 12.55 & \textbf{52.34} & 57.61 & \textbf{50.59} & 24.80 & 24.77 & 26.88 & 26.09 & 37.58 \\
loop4-CHASE (Ours) & skip & 0.4 & 2.04 & 12.30 & 51.85 & 60.21 & 49.33 & 26.80 & \textbf{25.36} & 27.39 & \textbf{27.31} & 38.32 \\
\midrule
\multicolumn{13}{l}{\textbf{370M Mamba-2 Hybrid}} \\
baseline & full & -- & 1.00 & 12.22 & \textbf{52.39} & 60.06 & \textbf{50.59} & 26.20 & 24.95 & 29.35 & 26.01 & 38.51 \\
loop4 & full & -- & 4.00 & \textbf{11.59} & 51.74 & \textbf{61.22} & 49.33 & \textbf{28.00} & \textbf{25.97} & 29.86 & \textbf{26.26} & \textbf{38.91} \\
loop4-Ouro & full & 0.8 & 4.00 & 12.12 & 51.09 & 39.45 & 50.04 & 27.40 & 25.84 & \textbf{30.29} & 26.09 & 35.74 \\
\bottomrule
\end{tabular}
}
\caption{Full downstream evaluation under the iso-parameter setting. \textbf{Exec.} is the average number of executed recurrent loops (\emph{full}$=4$; \emph{skip}$=$measured mean exit step). The exit-gated rows apply, to the loop4 model, the stock Ouro Stage-2 gate (\emph{loop4-Ouro}) and our CHASE gate (\emph{loop4-CHASE}); skip is defined for pure Mamba-2 only. Bold marks the best value within each model group.}
\label{tab:mamba2_exitgate_full}
\end{table*}

In addition to the iso-FLOPs comparison above, we evaluate looped models under an iso-parameter setting.
Here, the number of independently parameterized layers is fixed, while the number of recurrent loops is varied.
Thus, an $N\otimes R$ model has the same number of trainable parameters as its $N\otimes1$ baseline, but uses a larger effective computation depth at inference time.
This setting allows us to examine whether recurrent computation improves language modeling and downstream performance under a fixed parameter budget. Note that the 140M and 370M models use different backbone depths: the 140M models contain 24 parameterized layers, whereas the 370M models contain 48 parameterized layers.
Consequently, loop4 corresponds to an effective depth of 96 layers for 140M models and 192 layers for 370M models.
Therefore, the 140M and 370M results should be interpreted as separate iso-parameter comparisons within each model scale, rather than as directly depth-matched comparisons across scales.

Table~\ref{tab:mamba2_exitgate_full} reports the complete results for LLaMA-2 Transformer, Mamba-2, and Mamba-2 Hybrid backbones.
For the exit-gated rows we report both the stock Ouro gate (\emph{loop4-Ouro}) and our CHASE gate (\emph{loop4-CHASE}) at a representative threshold $q$.
Because the Ouro gate executes all four loops regardless of $q$ (it only relocates the readout), we report it as a full-compute reference; for CHASE, which skips the deeper loops, we choose the $q$ that attains the highest downstream average among the thresholds that reduce the executed loops below fixed 4-loop inference.
The full threshold sweep is reported in Appendix~\ref{app:exitgate_sweep}.

The iso-parameter results show that recurrent depth is generally useful, especially for validation perplexity.
For example, the 140M LLaMA-2 loop4 model improves Avg. from 37.79 to 38.41 over the non-looped baseline, and fixed-loop Mamba-2 variants also reduce PPL.
However, downstream performance is not monotonic with the number of loops.
The 140M Mamba-2 loop2 model improves PPL but suffers a large drop on BoolQ, and the loop4 model achieves the best PPL among fixed-loop variants while its downstream average remains close to the non-looped baseline.
This suggests that validation perplexity alone is insufficient to determine the optimal recurrent depth for downstream use.

The fixed-loop inference ablation in Table~\ref{tab:fixed_loop_ablation_140m} (Appendix~\ref{app:fixed_loop_ablation}) further shows that, under the final-loop pre-training objective, intermediate recurrent steps are poorly calibrated with respect to the language-modeling head: forcing fewer than $R$ loops at inference time yields severely degraded perplexity, while different downstream tasks prefer different effective depths within $r \le R$.
Naively truncating a fixed-loop model at inference time is therefore not a viable way to save computation, and no single forced depth is optimal for all tasks.
This motivates adaptive exit-state selection instead of always using the maximum recurrent depth.

We therefore attach a lightweight exit gate to pretrained fixed-loop checkpoints and continue training with the adaptive objectives described in Appendix~\ref{app:exitgate_training}.
At inference time, the gate predicts a distribution over recurrent exit steps for each token, and the threshold $q$ controls the compute--performance trade-off.
Smaller $q$ values encourage earlier exits, while larger values allocate more recurrent computation.
Thus, exit-gated inference can exploit intermediate recurrent states and avoid forcing all tokens to use the same fixed depth.

For the 140M Mamba-2 model, Table~\ref{tab:mamba2_exitgate_full} contrasts two inference modes. The stock Ouro gate (\emph{loop4-Ouro}) moves the readout to its predicted exit step but still executes all four recurrent loops, so it saves no computation. CHASE instead exits early and skips the deeper loops, roughly halving the executed loops: it reaches 38.82 Avg. at only 2.15 executed loops, essentially matching the full-compute Ouro gate (38.86 Avg. at 4 loops) and exceeding the fixed loop4 baseline (37.79).

The advantage grows at 370M. Here CHASE reaches 38.32 Avg. at 2.04 executed loops, surpassing the full-compute Ouro gate (37.58) and approaching the strongest fixed-depth model (loop4, 38.51) at about half its compute. Thus, even where a deeper fixed model keeps the peak accuracy, adaptive skipping recovers almost all of it at a fraction of the loops.

\begin{figure*}[t]
\centering
\includegraphics[width=\textwidth]{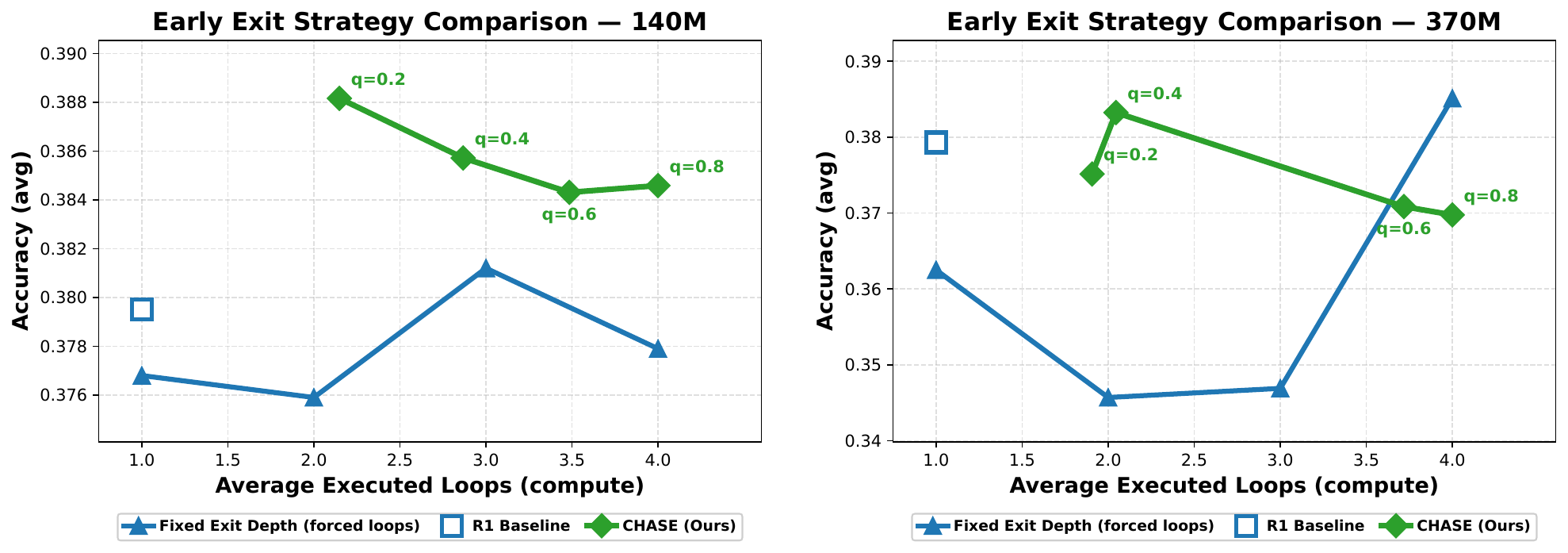}
\caption{Compute--performance trade-off under adaptive-depth inference (140M left, 370M right). The $x$-axis is the average number of \emph{executed} recurrent loops, i.e., the actual inference compute. \emph{Fixed Exit Depth} forces all tokens to a shared loop depth (the fixed-loop ablation of Tables~\ref{tab:fixed_loop_ablation_140m} and~\ref{tab:fixed_loop_ablation_370m}); \emph{CHASE} is our cache-hole--adapted gate swept over the threshold $q$, exiting early and skipping the deeper loops. At every compute budget the gated curve lies above the fixed-depth curve.}
\label{fig:exitgate_tradeoff}
\end{figure*}

Figure~\ref{fig:exitgate_tradeoff} summarizes the compute--performance trade-off for the 140M model in the left panel and the 370M model in the right panel. In both panels, CHASE traces a consistently better frontier than forcing every token to use a shared fixed exit depth: at a comparable number of executed loops, adaptive per-token exits retain substantially higher downstream accuracy. Naively forcing two loops yields only 37.59 and 34.57 Avg. at 140M and 370M, respectively, because intermediate recurrent states are miscalibrated with the language-modeling head. By contrast, the CHASE curve remains above the fixed-depth curve across the full compute range, and the separation persists at both model scales. This shows that the gain comes from allocating recurrent depth adaptively rather than simply using more computation. Executable skipping is possible only because the cache-hole adaptation (Section~\ref{sec:cache_hole_adaptation}) keeps early-exit recurrent states consistent under skipping; the stock gate, trained without this consistency, must execute every loop and can only relocate its readout.

Figure~\ref{fig:ppl_skip_compute} makes this failure concrete for the 370M models. As the gate skips more aggressively (fewer executed loops), the stock Ouro Stage-2 gate's validation perplexity explodes---by up to $+4788\%$ (from $12.2$ to nearly $600$) and erratically---because a skipped loop leaves the recurrent Mamba state without its update. CHASE instead stays within $+3$--$4\%$ of the full-compute perplexity across the whole range. Overall, recurrent depth is best treated as an adaptive computational resource, which CHASE converts into real, deployable compute savings without loss of downstream quality. These savings do materialize as wall-clock speedups, but only once the prefill is compute-bound: measured over batch size, sequence length, and model size, CHASE is no faster than full computation at batch~1, and reaches $1.8\times$ at batch~64, with larger models benefiting more than smaller ones (Section~\ref{sec:chase_inference}).

\subsection{Inference Cost of CHASE}
\label{sec:chase_inference}
Skipping deeper loops only pays off if the saved recurrent steps translate into wall-clock savings. Because tokens exit at different depths, the surviving tokens must be packed before each deeper loop, and the cost of a loop is set by the batch-maximum number of survivors rather than by the average exit depth. The average exit depth therefore \emph{overstates} the achievable speedup---the ragged-halting effect that motivates the fixed-iteration choice of \citet{deng2026lt2}. We measure the effect directly.

We benchmark prefill throughput and peak memory on a single RTX~5090 in bf16 on held-out validation documents, comparing \emph{full} (all $R{=}4$ loops) with \emph{skip} (CHASE) at matched thresholds. To isolate backbone packing, \emph{full} uses the same $O(1)$-in-loops readout policy as \emph{skip}, so the two differ only in whether deeper loops are executed for exited tokens.

Three factors govern the speedup. \textbf{Batch size} (Table~\ref{tab:chase_batch}): at batch~1 packing gives no benefit ($0.95$--$0.97\times$ at 140M) because the prefill is latency-bound, whereas at batch~64 CHASE reaches $1.73\times$ (140M) and $1.83\times$ (370M) at $q{=}0.2$. \textbf{Sequence length} (Table~\ref{tab:chase_length}): at $L\le512$ the speedup at $q{=}0.4$ is $\approx1.0\times$ and grows to $1.32\times$ at $L{=}3584$; the apparently large values at $q{=}0.2$ for short sequences are inflated because the \emph{full} baseline is itself overhead-throttled there, so only the compute-bound regime $L\ge1024$ should be read as a genuine gain. \textbf{Model size}: at matched batch and length the 370M model gains more than the 140M model ($1.59\times$ versus $1.47\times$ at $q{=}0.2$, batch~16), since a larger backbone spends a greater fraction of its time in the loops that packing removes. Peak memory is essentially identical for \emph{full} and \emph{skip}, since it is dominated by tensors that do not shrink with the number of executed loops; packing saves computation rather than memory.

Measured speedups remain below the ratio implied by the average exit depth---for example, $q{=}0.4$ at 370M exits after $2.76$ loops on average, implying $1.45\times$, while the measured value is $1.29\times$ at batch~16---which quantifies the ragged-halting gap.

\begin{table}[t]
\centering\footnotesize
\renewcommand{\arraystretch}{0.82}
\captionsetup{skip=3pt}
{\setlength{\tabcolsep}{3pt}
\begin{tabular}{lcccc@{\hspace{2pt}}c}
\toprule
\textbf{Batch} & \textbf{$q{=}0.2$} & \textbf{$q{=}0.4$} & \textbf{$q{=}0.6$} & \textbf{Mem.\ full} & \textbf{Mem.\ skip} \\
\midrule
\multicolumn{6}{l}{\textbf{140M Mamba-2}} \\
1 & 0.96$\times$ & 0.95$\times$ & 0.97$\times$ & 0.72 & 0.73 \\
4 & 0.85$\times$ & 0.87$\times$ & 0.88$\times$ & 1.92 & 1.94 \\
8 & 1.20$\times$ & 1.13$\times$ & 1.05$\times$ & 3.52 & 3.56 \\
16 & 1.47$\times$ & 1.17$\times$ & 1.09$\times$ & 6.72 & 6.81 \\
64 & 1.73$\times$ & 1.41$\times$ & 1.10$\times$ & 25.90 & 26.29 \\
\midrule
\multicolumn{6}{l}{\textbf{370M Mamba-2}} \\
4 & 1.29$\times$ & 1.10$\times$ & 1.04$\times$ & 2.41 & 2.43 \\
8 & 1.68$\times$ & 1.20$\times$ & 1.13$\times$ & 4.01 & 4.07 \\
16 & 1.59$\times$ & 1.29$\times$ & 1.11$\times$ & 7.23 & 7.34 \\
32 & 1.74$\times$ & 1.29$\times$ & 1.13$\times$ & 13.66 & 13.88 \\
64 & 1.83$\times$ & 1.34$\times$ & 1.13$\times$ & 26.52 & 27.05 \\
\bottomrule
\end{tabular}
}
\caption{CHASE speedup and peak memory versus batch size ($L{=}2048$, RTX~5090, bf16); memory is measured at $q{=}0.4$.}
\label{tab:chase_batch}
\vspace{0.25em}
\begin{tabular}{lcccc}
\toprule
\textbf{Length} & \textbf{$q{=}0.2$} & \textbf{$q{=}0.4$} & \textbf{$q{=}0.6$} & \textbf{Peak mem.\ (GB)} \\
\midrule
256 & 1.90$\times$ & 0.99$\times$ & 1.01$\times$ & 1.60 \\
512 & 1.87$\times$ & 1.00$\times$ & 1.01$\times$ & 2.41 \\
1024 & 1.58$\times$ & 1.29$\times$ & 1.06$\times$ & 4.01 \\
2048 & 1.59$\times$ & 1.29$\times$ & 1.13$\times$ & 7.23 \\
3584 & 1.76$\times$ & 1.32$\times$ & 1.10$\times$ & 12.05 \\
\bottomrule
\end{tabular}
\caption{CHASE speedup and peak memory versus sequence length (370M, batch size 16).}
\label{tab:chase_length}
\end{table}

\section{Conclusion}
We studied Looped Mamba, which reapplies a shared Mamba or Hybrid Mamba--Transformer backbone to gain effective depth without additional parameters. It improves over iso-parameter non-looped baselines on Mano and p-hop induction, and in pre-training it improves perplexity under a fixed parameter budget and stays competitive downstream, though deeper non-looped models remain ahead under strict iso-FLOPs comparisons. Turning adaptive depth into savings requires executing the exits, which fails without adaptation: skipped steps leave cache holes and perplexity diverges. CHASE removes this failure, so adaptive-depth inference matches or exceeds full-compute exit-state selection while executing about half of the recurrent steps. Recurrent depth is thus best treated not as a fixed maximum but as an adaptive computational resource.

\section*{Limitations}
This work has several limitations. 
First, our experiments are limited to models up to 370M parameters, and we have not yet validated the scaling behavior of Looped Mamba at billion-parameter or larger scales. 
Second, our language-model pre-training budget is 100B--150B tokens, which is modest compared with modern large-scale LLM training regimes. 
Thus, our results should be interpreted as controlled empirical evidence rather than a complete characterization of large-scale performance. 
Third, Mano and p-hop induction are synthetic reasoning tasks. 
Although they isolate compositional computation and recursive retrieval, they may not fully represent open-ended real-world reasoning. 
Fourth, our cache-hole adaptation (CHASE), and hence the executable skip inference it enables, is applied only to the pure Looped Mamba backbone: in the Mamba--Transformer hybrid the attention layers maintain a key--value cache, so skipping a recurrent step also leaves holes in this KV cache, which the present adaptation does not handle; we therefore report the hybrid models only with the full-compute Ouro gate and leave KV-aware cache-hole adaptation to future work. Finally, our Exit Gate experiments use fixed global thresholds $q$ to control inference-time computation, and do not yet optimize compute allocation per token, example, or task. 
Future work should evaluate larger models, longer training runs, more realistic reasoning benchmarks, more fine-grained adaptive-compute strategies, and cache-hole adaptation for hybrid attention key--value caches.

\bibliography{references}

\clearpage

\appendix

\section{Details of Two-Stage Exit-Gate Training}
\label{app:exitgate_training}

\paragraph{Exit-Gate Stage 1: adaptive exit-state selection.}
Following the entropy-regularized adaptive computation objective of Ouro~\citep{zhu2025scaling}, we attach an exit gate to the Looped Mamba model. Unlike the final-loop objective used in base pre-training, this stage uses the outputs from all recurrent steps. The exit gate operates at the token level: for token position $i$ and recurrent step $r$ with $r=1,\ldots,R-1$, it predicts an instantaneous exit probability
\[
\lambda_i^{(r)}
=
\sigma\!\left(\mathrm{Linear}_{\phi}(h_i^{(r)})\right)
\in (0,1),
\]
where $h_i^{(r)}$ is the hidden state of token $i$ after the $r$-th recurrent step, $\phi$ denotes the gate parameters, and $\sigma(\cdot)$ is the sigmoid function. The final recurrent step does not require a separate gate, since it receives all remaining probability mass.

For each token position, we define the survival probability as the probability of not exiting before a given recurrent step:
\[
S_i^{(0)}=1,
\qquad
S_i^{(r)}
=
\prod_{j=1}^{r}
\left(1-\lambda_i^{(j)}\right).
\]
The probability that token $i$ exits exactly at recurrent step $r$ is
\begin{equation}
\begin{split}
\pi_i^{(r)}
&= p_{\phi}(z_i=r\mid x) \\
&= \lambda_i^{(r)} S_i^{(r-1)},
\qquad r=1,\ldots,R-1.
\end{split}
\end{equation}
where $z_i$ denotes the latent exit step for token $i$. For the final recurrent step, we assign the remaining probability mass:
\[
\pi_i^{(R)}
=
p_{\phi}(z_i=R\mid x)
=
S_i^{(R-1)}.
\]
This defines a valid token-level distribution over recurrent steps:
\[
\sum_{r=1}^{R}\pi_i^{(r)}=1.
\]

Let
\[
\ell_i^{(r)}
=
-\log p_{\theta}^{(r)}(x_{i+1}\mid x_{\leq i})
\]
be the token-level next-token loss at recurrent step $r$. To avoid collapse to a single recurrent depth, we add an entropy regularizer over the token-level exit distribution:
\[
H_i
=
-
\sum_{r=1}^{R}
\pi_i^{(r)}
\log \pi_i^{(r)}.
\]
The Stage 1 objective is therefore
\[
\mathcal{L}_{\mathrm{stage1}}
=
\mathbb{E}_{x}
\left[
\frac{1}{M}
\sum_{i=1}^{M}
\left(
\sum_{r=1}^{R}
\pi_i^{(r)}
\ell_i^{(r)}
-
\beta H_i
\right)
\right],
\]
where $\beta$ controls the strength of entropy regularization, and $M$ denotes the number of valid next-token prediction positions. This objective encourages the gate to explore different recurrent depths while weighting each recurrent-step loss by the probability of exiting at that step.

At inference time, we use the cumulative exit probability
\[
\mathrm{CDF}_i(r\mid x)
=
\sum_{j=1}^{r}
\pi_i^{(j)}
\]
to determine the selected output step. Given a threshold $q\in[0,1]$, token $i$ selects recurrent step
\[
z_i(q)
=
\min
\left\{
r\in\{1,\ldots,R\}:
\mathrm{CDF}_i(r\mid x)\ge q
\right\}.
\]
The threshold $q$ controls the selected-depth--accuracy trade-off: smaller values favor earlier recurrent-step outputs, while larger values select later recurrent-step outputs.

\paragraph{Exit-Gate Stage 2: focused exit-gate training.}
Following the focused adaptive gate training stage of Ouro~\citep{zhu2025scaling}, we further calibrate the exit gate after entropy-regularized training. In this stage, the Mamba backbone and the language-modeling head are frozen, and only the exit-gate parameters $\phi$ are updated. The goal is to train the gate to decide whether continuing to the next recurrent step is useful, based on the realized improvement in next-token loss.

We first detach the per-step token losses from the computation graph:
\[
\bar{\ell}_i^{(r)}
=
\mathrm{sg}\!\left(\ell_i^{(r)}\right),
\]
where $\mathrm{sg}(\cdot)$ denotes the stop-gradient operator. For each non-final recurrent step $r\in\{1,\ldots,R-1\}$, we define the improvement obtained by continuing from step $r$ to step $r+1$ as
\[
I_i^{(r)}
=
\max
\left(
0,\,
\bar{\ell}_i^{(r)}
-
\bar{\ell}_i^{(r+1)}
\right).
\]
A larger $I_i^{(r)}$ means that the additional recurrent step reduces the loss, suggesting that the model should continue rather than exit.

We convert this improvement into a soft continuation target:
\[
w_i^{(r)}
=
\sigma\!\left(
\kappa
\left(
I_i^{(r)}-\gamma
\right)
\right),
\]
where $\kappa$ controls the sharpness of the target and $\gamma$ is the minimum improvement threshold. When the improvement is larger than $\gamma$, $w_i^{(r)}$ is close to 1 and the gate is encouraged to continue; otherwise, it is encouraged to exit.

Since $\lambda_i^{(r)}$ is the predicted exit probability, the predicted continuation probability is $1-\lambda_i^{(r)}$. We train the gate by matching this predicted continuation probability to $w_i^{(r)}$ using binary cross-entropy:
\begin{equation}
\begin{split}
\mathcal{L}_{\mathrm{adapt}}^{(r)}
= -\frac{1}{M} \sum_{i=1}^{M} \Big[
& w_i^{(r)} \log\!\left(1-\lambda_i^{(r)}\right) \\
& + \left(1-w_i^{(r)}\right) \log \lambda_i^{(r)} \Big].
\end{split}
\end{equation}

The total Stage 2 objective averages over all non-final recurrent steps:
\[
\mathcal{L}_{\mathrm{stage2}}
=
\frac{1}{R-1}
\sum_{r=1}^{R-1}
\mathcal{L}_{\mathrm{adapt}}^{(r)}.
\]
This objective penalizes early exits when another recurrent step would substantially reduce the loss, while also penalizing unnecessary continuation when the additional step provides little improvement. At inference time, the calibrated gate is used with the same cumulative-probability threshold rule as in Stage 1.

\section{Mano: Knowledge Manipulation}
\label{app:mano}
\paragraph{Dataset.}We use the Mano task to evaluate the model's ability to manipulate knowledge stored in its parameters without explicit intermediate reasoning. The dataset consists of arithmetic expressions over the finite field $\mathbb{F}_{23}$. Each instance corresponds to a tree-structured expression with $\ell$ binary operations, where $\ell \leq L$ and $L$ denotes the maximum expression length. During training, $\ell$ is uniformly sampled from $[1,L]$; during evaluation, we use only the hardest expressions with $\ell=L$. The expressions are presented in prefix notation. A length-3 instance is written as
\[
\texttt{<bos> <len\_3> - * a b + c d <ans> ans}.
\]
All operations are performed in $\mathbb{F}_{23}$, and the task uses only the operators $\{+, -, \ast\}$. The vocabulary consists of the numbers $0,\ldots,22$, the three operators, the special tokens $\texttt{<bos>}$, $\texttt{<ans>}$, $\texttt{<eos>}$, and $\texttt{<pad>}$, and length tokens $\texttt{<len\_i>}$ for $i\in[0,L]$.

\paragraph{Model.}We replace the Transformer blocks in the looped-model architecture with Mamba-2 blocks and study recurrent parameter sharing in this architecture. A model is denoted by $N \otimes R$, where $N$ is the number of independently parameterized Mamba layers and $R$ is the number of recurrent steps. Thus, $R=1$ corresponds to a standard non-looped Mamba model, while $R>1$ corresponds to a Looped Mamba model. All models use hidden dimension 1024. Training and evaluation are performed using only the output from the final recurrent step. Since our Mamba implementation disables the feed-forward sublayer, we use two consecutive Mamba-2 mixer blocks as a proxy for the two-sub-layer structure of a Transformer block, namely attention followed by an MLP. This design matches the number of residual transformations at a coarse level, but should not be interpreted as an exact architectural equivalence. We therefore report results under explicit effective-depth, iso-parameter, and iso-FLOPs comparisons.

\paragraph{Training details.}
We use the AdamW optimizer with $\beta_1,\beta_2=0.9,0.98$, $\epsilon=10^{-6}$, weight decay 0.1, and gradient clipping with maximum norm 1.0. We employ 1000 warmup steps followed by a cosine learning-rate schedule whose minimum learning rate is 0.1 of the peak learning rate. Training uses bf16, sequence packing, context length 1024, and global batch size 128. For the three difficulty levels, we train for 80K, 110K, and 200K steps, respectively. The corresponding peak learning rates are $5\cdot10^{-4}$, $1\cdot10^{-4}$, and $5\cdot10^{-5}$. Training samples are generated online, and multiple Mano instances are packed into each 1024-token chunk. For each configuration, we train the model three times with different random seeds and report the run with the maximum average evaluation accuracy on the test set. We adopt this best-of-three reporting protocol because our goal is to study the expressivity power of the model under a given configuration, rather than to characterize average training stability across random seeds.

\paragraph{Evaluation.}During evaluation, each sample contains a single expression of length $\ell=L$, without packing. Since the final answer is a single token in $\mathbb{F}_{23}$, we report exact-match accuracy at the answer position. Specifically, the model predicts the single answer token following $\texttt{<ans>}$, and the prediction is counted as correct if it matches the ground-truth answer. For each difficulty level, we evaluate on 2000 deterministic samples generated with a fixed random seed.

\section{p-hop Induction}
\label{app:phop}

\paragraph{Dataset.}We use the p-hop induction task to evaluate recursive contextual retrieval. Given a sequence $v=(v_1,\ldots,v_n)$ of length $n=256$, each token is sampled from an alphabet $\Sigma$ of size 4. A 1-hop induction step starts from the final position, finds a previous occurrence of the target token, and returns the token immediately following that occurrence. The p-hop task recursively applies this operation $p$ times. We evaluate two difficulty levels, $p\in\{16,32\}$. To ensure that every instance has a valid answer, our implementation uses a chain-first construction: it first samples a valid hop chain of length $p$, fills the remaining positions subject to constraints, and then verifies that the actual hop trace exactly matches the constructed chain.

\paragraph{Model.} We use the same Looped Mamba architecture as in the Mano experiments. A model is denoted by $N \otimes R$, where $N$ is the number of independently parameterized Mamba-2 layers and $R$ is the number of recurrent steps. Therefore, the number of parameters scales with $N$, while FLOPs scale with the effective depth $NR$. All p-hop experiments use hidden dimension $d=256$. The final recurrent step is used for both training loss and evaluation. 

\paragraph{Training details.}We train with AdamW, weight decay 0.1, and maximum gradient norm 1.0. We use 2000 warmup steps followed by a cosine learning-rate schedule. The peak learning rate is $2\times10^{-4}$, and the minimum learning rate is $2\times10^{-5}$, i.e., 0.1 of the peak value. Training uses bf16. The context length is set to 272. Since each p-hop sample has length $1+256+3=260$, this setting effectively corresponds to using one p-hop instance per training sequence rather than packing multiple instances into a longer context. The loss is applied only to the answer token position. For each configuration, we train the model three times with different random seeds and report the run with the maximum average evaluation accuracy on the test set. We adopt this best-of-three reporting protocol because our goal is to study the expressivity power of the model under a given configuration, rather than to characterize average training stability across random seeds.

\paragraph{Evaluation.}The evaluation set contains 10,000 deterministic samples generated with a fixed random seed. During evaluation, each sample is fed independently into the model, and we compute exact-match accuracy at the answer position after $\texttt{<ans>}$. Specifically, the prediction is the argmax of the logits at the answer position, and it is counted as correct if it matches the ground-truth answer $y$. Since the answer is one of four alphabet tokens, random guessing gives approximately 25\% accuracy. The corresponding exact-match results are reported in Table~\ref{tab:phop_mamba} in Section~\ref{sec:phop}.

\section{Language Model Training Details}
\label{app:training_details}

\subsection{Common Language Model Pre-training Details}
\label{app:lm_pretraining_details}

\paragraph{Overall procedure.}
Our language-model experiments are organized into three steps. First, we pre-train fixed-loop language models without exit gates. This pre-training setting is shared by the LLaMA-2 Transformer, Mamba-2, and Mamba-2 Hybrid models described in Sections~\ref{app:llama2_details}, \ref{app:mamba2_details}, and \ref{app:mamba2_hybrid_details}, respectively. Second, we evaluate the pretrained fixed-loop models under the corresponding architecture and loop configurations. Third, for the exit-gate experiments, we initialize from a pretrained fixed-loop checkpoint and continue training with the exit gate enabled, as described in Section~\ref{app:exitgate_details}.

\paragraph{Pre-training configuration.}
Unless otherwise specified, the 140M models use $d_{\mathrm{model}}=768$ and 24 layers. Larger 370M models modify only the architecture-specific dimensions and depth, as listed in the corresponding architecture tables. The shared pre-training hyperparameters are summarized in Table~\ref{tab:app_lm_pretraining_hparams}.

\begin{table}[t]
\centering
\begin{tabular}{ll}
\toprule
\textbf{Parameter} & \textbf{Value} \\
\midrule
Context length & 2048 \\
Vocabulary size & 32,000 \\
Optimizer & AdamW \\
AdamW betas & $(0.9, 0.95)$ \\
Weight decay & 0.1 \\
Learning rate schedule & cosine \\
Maximum scheduler LR & $8.0 \times 10^{-4}$ \\
Minimum scheduler LR & $8.0 \times 10^{-5}$ \\
Warmup ratio & 0.1 \\
Gradient clipping & 1.0 \\
Precision & bfloat16 \\
Random seed & 3407 \\
\bottomrule
\end{tabular}
\caption{Common pre-training hyperparameters for the fixed-loop language models. The LLaMA-2 Transformer, Mamba-2, and Mamba-2 Hybrid models are first pretrained under this shared setting before any exit-gate continued training. }
\label{tab:app_lm_pretraining_hparams}
\end{table}
\subsection{LLaMA-2 Transformer Details}
\label{app:llama2_details}

\paragraph{Architecture.}
For the LLaMA-2 Transformer experiments, we use the Transformer ablation branch of our looped language model implementation. The model replaces the Mamba-2 mixer with LLaMA-style causal self-attention while keeping the same looped backbone interface. Unless otherwise specified, the Transformer models use a model dimension of $d_{\mathrm{model}}=768$. To make the Transformer baseline comparable to the Mamba-2 model with $d_{\mathrm{model}}=768$ under a similar parameter budget, we set the gated MLP intermediate dimension to $d_{\mathrm{intermediate}}=640$. This intermediate dimension is not intended to follow the standard LLaMA MLP expansion ratio; instead, it is chosen as a parameter-matching configuration for comparison with Mamba-2. The configuration is summarized in Table~\ref{tab:app_llama2_config}.

\begin{table}[t]
\centering
\begin{tabular}{ll}
\toprule
\textbf{Parameter} & \textbf{Value} \\
\midrule
Architecture & LLaMA-2 \\
$d_{\mathrm{model}}$ & 768 \\
$d_{\mathrm{intermediate}}$ & 640 \\
Layers & 6, 12, or 24 \\ 
Attention head dimension & 64 \\
Number of attention heads & 12 \\
Position embedding & RoPE \\
Normalization & RMSNorm \\
Fused add-norm & yes \\
Residual in FP32 & yes \\
Tied embeddings & no \\
Context length & 2048 \\
Vocabulary size & 32,000 \\
\bottomrule
\end{tabular}
\caption{LLaMA-2 Transformer architecture configuration.}
\label{tab:app_llama2_config}
\end{table}

\subsection{Mamba-2 Details}
\label{app:mamba2_details}

\paragraph{Architecture.}
For the Mamba-2 experiments, we use the same looped language model interface as the LLaMA-2 Transformer baseline, but replace the Transformer self-attention mixer with a Mamba-2 mixer. The 140M Mamba-2 model uses $d_{\mathrm{model}}=768$ with 24 layers, while the 370M model increases the model dimension to $d_{\mathrm{model}}=1024$ and uses 48 layers. For both sizes, we set the external MLP intermediate dimension to $d_{\mathrm{intermediate}}=0$, since the Mamba-2 block already contains its own sequence mixing and channel projection components. The configuration is summarized in Table~\ref{tab:app_mamba2_config}.

\begin{table}[t]
\centering
\begin{tabular}{lll}
\toprule
\textbf{Parameter} & \textbf{140M} & \textbf{370M} \\
\midrule
Architecture & Mamba-2 & Mamba-2 \\
$d_{\mathrm{model}}$ & 768 & 1024 \\
Number of layers & 24 & 48 \\
External MLP & no & no \\
Attention layers & no & no \\
Normalization & RMSNorm & RMSNorm \\
Fused add-norm & yes & yes \\
Residual in FP32 & yes & yes \\
Tied embeddings & no & no \\
Context length & 2048 & 2048 \\
Vocabulary size & 32,000 & 32,000 \\
\bottomrule
\end{tabular}
\caption{Mamba-2 architecture configurations.}
\label{tab:app_mamba2_config}
\end{table}

\subsection{Mamba-2 Hybrid Details}
\label{app:mamba2_hybrid_details}

\paragraph{Architecture.}
For the Mamba-2 Hybrid experiments, we use the same looped language model interface as the Mamba-2 models, but replace a small number of Mamba-2 layers with causal attention layers. The attention layers are specified by the 0-based indices in \texttt{attn\_layer\_idx}; all remaining layers use the Mamba-2 mixer. The detailed configurations are summarized in Table~\ref{tab:app_mamba2_hybrid_config}. The hybrid models do not use learned position embeddings, while the inserted attention layers use RoPE~\citep{su2023roformerenhancedtransformerrotary} with rotary dimension 64. 
\begin{table}[t]
\centering
\setlength{\tabcolsep}{4pt}
\begin{tabular}{lcc}
\toprule
& \textbf{140M} & \textbf{370M} \\
\midrule
$d_{\mathrm{model}}$            & 768            & 1024           \\
Layers                          & 24             & 48             \\
Attention indices         & $[5,10,15,20]$ & $[6,18,30,42]$ \\
Attention heads                 & $12 \times 64$ & $16 \times 64$ \\
RoPE dim.                       & 64             & 64             \\
\bottomrule
\end{tabular}
\caption{Mamba-2 Hybrid architecture configurations. Attention layer indices are 0-based. All non-attention layers are Mamba-2 layers. The models use no external MLP and no learned position embedding; RoPE is used only in the inserted attention layers.}
\label{tab:app_mamba2_hybrid_config}
\end{table}

\subsection{Continued Training: Exit-Gate Stage-1 Training Details}
\label{app:exitgate_details}

\paragraph{Training procedure.}
For the exit-gate experiments, we initialize from a pretrained fixed-loop checkpoint and continue training with the exit gate enabled. The backbone architecture is kept unchanged, while a lightweight linear gate is added on top of the hidden states to predict whether each token should exit at each recurrent loop. The exit-gate bias is initialized to $-2.0$, giving a conservative initial exit probability. The configuration is summarized in Table~\ref{tab:app_exitgate_training}.

\paragraph{Entropy coefficient $\beta$ ablation.}
We conduct a stage-1 ablation on the entropy coefficient $\beta$ for the exit-gate continuation training. In this ablation, we focus on the loop-4 setting with the maximum number of recurrent loops $R=4$. For each token, the exit gate defines an exit distribution over recurrent loops:
\[
p_{i,r}
=
\lambda_{i,r}
\prod_{j<r}(1-\lambda_{i,j}),
\qquad
\lambda_{i,r}=\sigma(g(h_{i,r})),
\]
for $r<R$, and the remaining probability mass is assigned to the final loop:
\[
p_{i,R}
=
\prod_{j<R}(1-\lambda_{i,j}).
\]
The exit entropy for token $i$ is computed as
\[
H_i
=
-\sum_{r=1}^{R} p_{i,r}\log(p_{i,r}+\epsilon),
\]
and the exit entropy is averaged over valid tokens:
\[
\bar{H}_{\mathrm{exit}}
=
\frac{1}{|\mathcal{V}|}
\sum_{i\in\mathcal{V}} H_i,
\]
where $\mathcal{V}$ denotes the set of valid non-padding tokens. For $R=4$, the maximum possible entropy is
\[
H_{\max}=\log R=\log 4 \approx 1.386,
\]
which is achieved when the exit distribution is uniform over the four loops. In practice, however, a fully uniform exit distribution is not ideal because it does not sufficiently favor later, more accurate loops. We therefore regard an exit entropy around $1.0$ as a desirable regime: the distribution remains non-collapsed, while still allocating more probability mass to later recurrent loops.
We also monitor the expected selected exit step. For a token $i$, this is
\[
s_i
=
\sum_{r=1}^{R} r\,p_{i,r}.
\]
The logged metric is the average of this quantity over valid tokens:
\[
\bar{s}
=
\frac{1}{|\mathcal{V}|}
\sum_{i\in\mathcal{V}} s_i,
\]
For the loop-4 exit-gate models, we find that an average expected step around $3.2$ is a good operating point, since it provides adaptive exit-state selection while favoring later recurrent-step outputs. The continuation stage is sensitive to the entropy coefficient $\beta$: too small a value can lead to a collapsed exit distribution, whereas too large a value can over-regularize the exit behavior. Based on the stage-1 ablation, we use $\beta=0.25$ as the final condition for the exit-gate continuation experiments.

\begin{table}[t]
\centering
\begin{tabular}{ll}
\toprule
\textbf{Parameter} & \textbf{Value} \\
\midrule
Exit-gate bias initialization & $-2.0$ \\
Entropy coefficient $\beta$ & 0.25 \\
Learning rate schedule & cosine \\
Maximum learning rate & $1.5 \times 10^{-4}$ \\
Minimum learning rate & $8.0 \times 10^{-5}$ \\
Warmup ratio & 0.002 \\
Optimizer & AdamW \\
AdamW betas & $(0.9, 0.95)$ \\
Weight decay & 0.1 \\
Gradient clipping & 0.6 \\
\bottomrule
\end{tabular}
\caption{Exit-gate continuation training configuration. The exit-gate models are initialized from pretrained fixed-loop checkpoints and trained with an expected loop loss plus entropy regularization.}
\label{tab:app_exitgate_training}
\end{table}

\begin{figure}[t]
\centering
\includegraphics[width=0.72\linewidth]{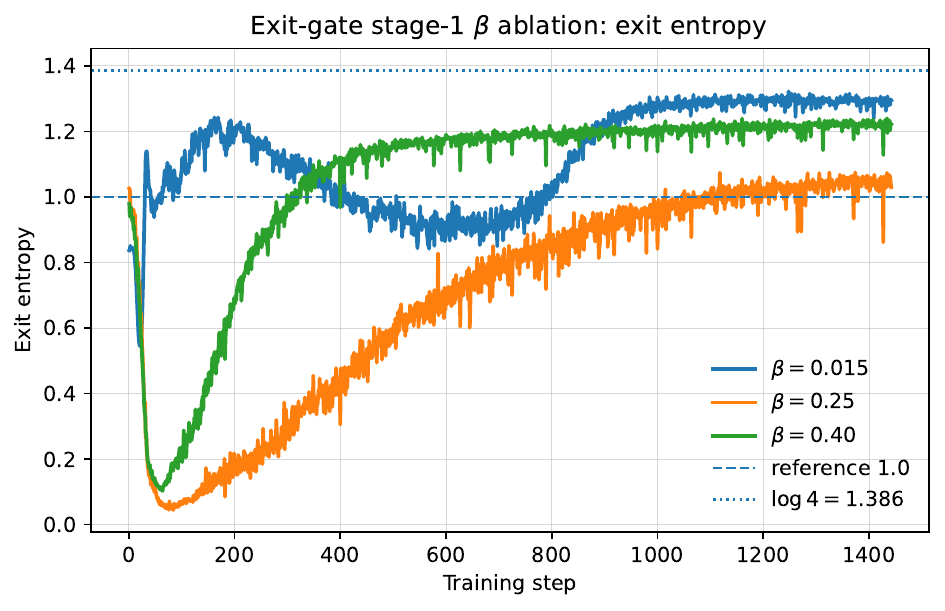}
\caption{Exit entropy during stage-1 exit-gate continuation training for different entropy coefficients $\beta$. All curves are truncated to the common training range. For $R=4$, the maximum entropy is $\log 4 \approx 1.386$, and we use an entropy around $1.0$ as a practical reference for a non-collapsed exit distribution.}
\label{fig:app_exitgate_beta_entropy}
\end{figure}

\begin{figure}[t]
\centering
\includegraphics[width=0.72\linewidth]{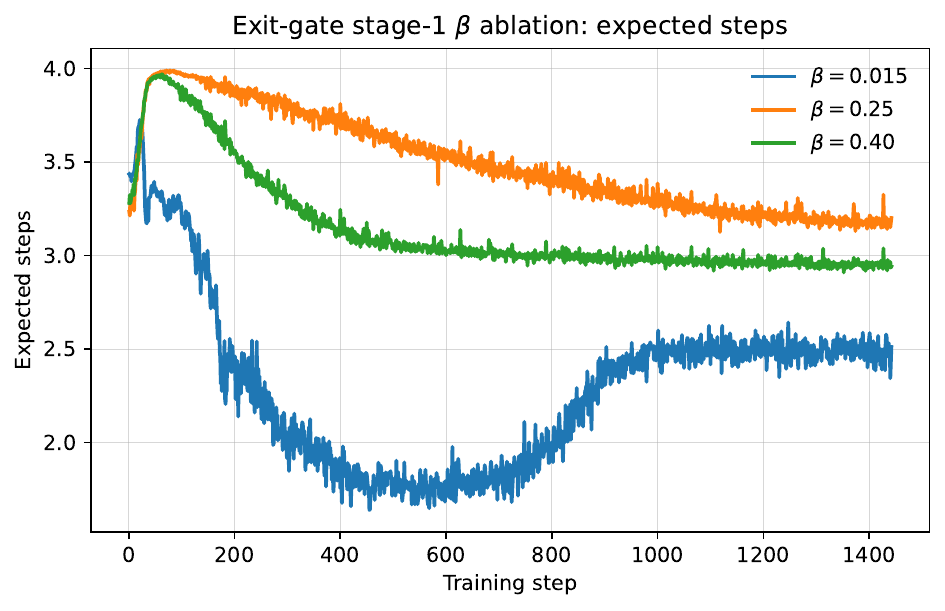}
\caption{Expected selected exit step during stage-1 exit-gate continuation training for different entropy coefficients $\beta$. We use this metric as a diagnostic for the prediction depth selected by the exit gate.}
\label{fig:app_exitgate_beta_expected_steps}
\end{figure}

\subsection{Continued Training: Cache-Hole Adaptation (CHASE)}
\label{app:chase_training}
Starting from the Stage-1 exit-gate weights, cache-hole adaptation (CHASE) continues training with \emph{holed} forward passes, so that the recurrent Mamba state stays consistent when deeper loops are skipped at inference. We use the depth-space hole sampler: for each holed position a target exit depth is drawn uniformly from $[1.85,3.45]$, discretized into $8$ bins, over the $R{=}4$ recurrent steps, and the state update is dropped for all steps beyond the sampled depth; a fraction $\mathrm{hp}{=}0.5$ of positions are holed. The training objective is the original entropy-regularized Stage-1 loss (the expected per-step next-token loss minus $\beta$ times the entropy of the exit distribution, with $\beta{=}0.25$), evaluated on the holed forward pass. We warm-start from the Stage-1 weights and use a peak learning rate of $1.5\times10^{-4}$ with a cosine schedule; all other optimizer settings match the Stage-1 configuration. The 140M model is adapted for 800 iterations and the 370M model for 2{,}000 iterations. The subsequent Exit-Gate Stage-2 gate calibration is then run on the frozen adapted backbone.

\subsection{Continued Training: Exit-Gate Stage-2 Training Details}
\label{app:exitgate_stage2_details}
\paragraph{Training procedure.}
In Stage-2, we further refine the exit gate after the Stage-1 continuation training. We initialize the model from the Stage-1 exit-gate checkpoint and keep the architecture unchanged. Unlike Stage-1, all parameters except the exit gate are frozen, including the token embedding, recurrent backbone, final normalization, and language modeling head. Thus, Stage-2 only updates the lightweight exit-gate parameters. The configuration is summarized in Table~\ref{tab:app_exitgate_stage2_training}.

\begin{table}[t]
\centering
\begin{tabular}{ll}
\toprule
\textbf{Parameter} & \textbf{Value} \\
\midrule
Improvement threshold $\gamma$ & 0.005 \\
Sharpness $k$ & 50 \\
Learning rate schedule & cosine \\
Maximum learning rate & $1.5 \times 10^{-4}$ \\
Minimum learning rate & $1.5 \times 10^{-5}$ \\
Warmup ratio & 0.01 \\
Optimizer & AdamW \\
AdamW betas & $(0.9, 0.95)$ \\
Weight decay & 0.1 \\
Gradient clipping & 1.0 \\
\bottomrule
\end{tabular}
\caption{Exit-gate Stage-2 continued training configuration. Stage-2 freezes the pretrained language model and refines only the exit gate using an adaptive continuation objective.}
\label{tab:app_exitgate_stage2_training}
\end{table}

\section{Fixed-Loop Inference Ablation}
\label{app:fixed_loop_ablation}
Tables~\ref{tab:fixed_loop_ablation_140m}--\ref{tab:fixed_loop_ablation_370m_hybrid} report the fixed-loop inference ablation results for the 140M and 370M Mamba-2 and Mamba-2 Hybrid looped models. Each model is trained as a 4-loop model without an exit gate and evaluated by forcing $r\in\{1,2,3,4\}$ recurrent loops at inference time. PPL denotes validation perplexity, and Avg. denotes the average score over the seven downstream tasks shown in the tables. Across all models, forcing fewer than four loops yields severely degraded perplexity, indicating that intermediate recurrent steps are not calibrated with the language-modeling head under the final-loop pre-training objective; this motivates the gate-based adaptive-depth training described in Appendix~\ref{app:exitgate_training}.

\begin{table*}[t]
\centering
\scriptsize
\caption{Fixed-loop inference ablation for the 140M Mamba-2 looped model. The model is trained as a 4-loop model without an exit gate, and evaluated by forcing different numbers of recurrent loops at inference time. Avg. denotes the average score over the seven downstream tasks shown in the table.}
\label{tab:fixed_loop_ablation_140m}
\resizebox{\textwidth}{!}{
\begin{tabular}{cccccccccc}
\toprule
\textbf{Forced Loops}
& \textbf{PPL}
& \textbf{PIQA}
& \textbf{BoolQ}
& \textbf{WG}
& \textbf{OBQA}
& \textbf{HS}
& \textbf{ARC-C}
& \textbf{ARC-E}
& \textbf{Avg.} \\
\midrule
1 & 46374.62 & 52.34 & 61.96 & 47.67 & 25.20 & 25.49 & 27.05 & 24.03 & 37.68 \\
2 & 12906.86 & 51.41 & 61.93 & 47.59 & 25.00 & 25.22 & 28.24 & 23.74 & 37.59 \\
3 & 464.82   & 52.34 & 61.41 & 49.49 & 25.60 & 25.92 & 27.13 & 24.92 & 38.12 \\
4 & 14.68    & 52.34 & 58.87 & 48.22 & 27.00 & 25.21 & 28.07 & 24.83 & 37.79 \\
\bottomrule
\end{tabular}
}
\end{table*}

\begin{table*}[t]
\centering
\scriptsize
\caption{Fixed-loop inference ablation for the 140M Mamba-2 Hybrid looped model. The model is trained as a 4-loop model without an exit gate, and evaluated by forcing different numbers of recurrent loops at inference time. Avg. denotes the average score over the seven downstream tasks shown in the table.}
\label{tab:fixed_loop_ablation_140m_hybrid}
\resizebox{\textwidth}{!}{
\begin{tabular}{cccccccccc}
\toprule
\textbf{Forced Loops}
& \textbf{PPL}
& \textbf{PIQA}
& \textbf{BoolQ}
& \textbf{WG}
& \textbf{OBQA}
& \textbf{HS}
& \textbf{ARC-C}
& \textbf{ARC-E}
& \textbf{Avg.} \\
\midrule
1 & 24061.87 & 53.21 & 37.83 & 50.20 & 25.80 & 26.06 & 27.73 & 26.01 & 35.26 \\
2 & 12304.00 & 52.18 & 37.83 & 51.22 & 25.80 & 25.62 & 29.01 & 26.14 & 35.40 \\
3 & 508.38   & 52.61 & 37.83 & 50.67 & 25.80 & 25.68 & 27.65 & 25.80 & 35.15 \\
4 & 14.37    & 51.31 & 45.87 & 47.43 & 26.80 & 25.07 & 28.92 & 26.09 & 35.93 \\
\bottomrule
\end{tabular}
}
\end{table*}

\begin{table*}[t]
\centering
\scriptsize
\caption{Fixed-loop inference ablation for the 370M Mamba-2 looped model. The model is trained as a 4-loop model without an exit gate, and evaluated by forcing different numbers of recurrent loops at inference time. Avg. denotes the average score over the seven downstream tasks shown in the table.}
\label{tab:fixed_loop_ablation_370m}
\resizebox{\textwidth}{!}{
\begin{tabular}{cccccccccc}
\toprule
\textbf{Forced Loops}
& \textbf{PPL}
& \textbf{PIQA}
& \textbf{BoolQ}
& \textbf{WG}
& \textbf{OBQA}
& \textbf{HS}
& \textbf{ARC-C}
& \textbf{ARC-E}
& \textbf{Avg.} \\
\midrule
1 & 1933.70 & 50.92 & 48.99 & 49.64 & 26.00 & 24.16 & 27.90 & 26.14 & 36.25 \\
2 & 1906.49 & 51.69 & 37.83 & 49.49 & 27.20 & 21.97 & 28.41 & 25.42 & 34.57 \\
3 & 832.53  & 51.09 & 38.04 & 50.43 & 25.40 & 23.07 & 28.75 & 26.05 & 34.69 \\
4 & 12.06   & 51.90 & 61.96 & 48.15 & 27.80 & 24.72 & 29.10 & 25.93 & 38.51 \\
\bottomrule
\end{tabular}
}
\end{table*}
\begin{table*}[t]
\centering
\scriptsize
\caption{Fixed-loop inference ablation for the 370M Mamba-2 Hybrid looped model. The model is trained as a 4-loop model without an exit gate, and evaluated by forcing different numbers of recurrent loops at inference time. Avg. denotes the average score over the seven downstream tasks shown in the table.}
\label{tab:fixed_loop_ablation_370m_hybrid}
\resizebox{\textwidth}{!}{
\begin{tabular}{cccccccccc}
\toprule
\textbf{Forced Loops}
& \textbf{PPL}
& \textbf{PIQA}
& \textbf{BoolQ}
& \textbf{WG}
& \textbf{OBQA}
& \textbf{HS}
& \textbf{ARC-C}
& \textbf{ARC-E}
& \textbf{Avg.} \\
\midrule
1 & 3282.60 & 51.47 & 60.80 & 49.01 & 26.00 & 24.54 & 26.62 & 25.17 & 37.66 \\
2 & 4692.78 & 49.89 & 44.31 & 50.04 & 25.60 & 21.27 & 27.30 & 24.83 & 34.75 \\
3 & 1660.17 & 51.41 & 37.77 & 51.14 & 25.40 & 23.23 & 27.82 & 25.04 & 34.54 \\
4 & 11.59   & 51.74 & 61.22 & 49.33 & 28.00 & 25.97 & 29.86 & 26.26 & \textbf{38.91} \\
\bottomrule
\end{tabular}
}
\end{table*}

\section{Exit-Gate Threshold Sweeps}
\label{app:exitgate_sweep}
We provide the full exit-gate threshold sweep results for the four looped models studied in the main text. 
Starting from fixed-loop checkpoints, we attach a lightweight exit gate and continue training the models with the adaptive-depth objectives described in Appendix~\ref{app:exitgate_training}. 
We report results for both Stage~1 and Stage~2 exit-gate training. 
For each stage, we evaluate inference with threshold values $q \in \{0.2, 0.4, 0.6, 0.8, 1.0\}$, where smaller thresholds select earlier recurrent-step outputs, whereas larger thresholds select later recurrent-step outputs.

Tables~\ref{tab:exitgate_q_sweep_140m}, \ref{tab:exitgate_q_sweep_140m_hybrid}, \ref{tab:exitgate_q_sweep_370m}, and \ref{tab:exitgate_q_sweep_370m_hybrid} report the downstream performance of the 140M Mamba-2, 140M Mamba-2 Hybrid, 370M Mamba-2, and 370M Mamba-2 Hybrid models, respectively. 
These sweeps illustrate how the exit threshold controls the selected-depth--performance trade-off after gate training, and show that the best threshold is not universal across model scales, architectures, or downstream tasks. 
As in the fixed-loop ablation, different tasks can prefer different effective recurrent depths, which motivates adaptive exit instead of using a single fixed loop count for all inputs.

\begin{table*}[t]
\centering
\caption{Threshold sweep for the 140M Mamba-2 exit-gated model. We report Stage 1 and Stage 2 exit-gate models with $q\in\{0.2,0.4,0.6,0.8,1.0\}$. Iter denotes the average exit step selected by the gate.}
\label{tab:exitgate_q_sweep_140m}
\resizebox{\textwidth}{!}{
\begin{tabular}{lccccccccccc}
\toprule
\textbf{Stage}
& \textbf{$q$}
& \textbf{Iter}
& \textbf{PPL}
& \textbf{PIQA}
& \textbf{BoolQ}
& \textbf{WG}
& \textbf{OBQA}
& \textbf{HS}
& \textbf{ARC-C}
& \textbf{ARC-E}
& \textbf{Avg.} \\
\midrule
Stage 1 & 1.00 & 4.00 & 15.54 & 52.18 & 61.59 & 50.51 & 26.80 & 24.60 & 28.07 & 25.00 & 38.39 \\
Stage 1 & 0.80 & 4.00 & 15.54 & 52.18 & 61.59 & 50.04 & 26.60 & 25.27 & 27.90 & 25.17 & 38.39 \\
Stage 1 & 0.60 & 3.43 & 15.54 & 51.47 & 62.23 & 47.75 & 26.80 & 24.67 & 29.78 & 25.04 & 38.25 \\
Stage 1 & 0.40 & 2.82 & 15.57 & 52.50 & 62.23 & 50.67 & 24.60 & 26.12 & 28.84 & 25.04 & 38.57 \\
Stage 1 & 0.20 & 2.38 & 16.22 & 52.72 & 62.17 & 47.04 & 24.40 & 26.14 & 28.50 & 24.75 & 37.96 \\
\midrule
Stage 2 & 1.00 & 4.00 & 15.54 & 52.18 & 61.59 & 49.88 & 26.60 & 25.32 & 27.90 & 25.17 & 38.38 \\
Stage 2 & 0.80 & 3.00 & 15.54 & 52.99 & 62.23 & 50.20 & 26.00 & 26.01 & 28.92 & 25.72 & \textbf{38.87} \\
Stage 2 & 0.60 & 2.63 & 15.78 & 53.05 & 62.17 & 49.09 & 24.60 & 26.36 & 28.33 & 25.21 & 38.40 \\
Stage 2 & 0.40 & 2.27 & 16.50 & 52.61 & 62.17 & 48.86 & 24.20 & 26.11 & 28.58 & 24.71 & 38.18 \\
Stage 2 & 0.20 & 1.92 & 18.14 & 52.39 & 62.17 & 50.20 & 26.40 & 26.34 & 27.47 & 24.41 & 38.48 \\
\bottomrule
\end{tabular}
}
\end{table*}

\begin{table*}[t]
\centering
\scriptsize
\caption{Threshold sweep for the 140M Mamba-2 Hybrid exit-gated model. We report Stage 1 and Stage 2 exit-gate models with $q\in\{0.2,0.4,0.6,0.8,1.0\}$. Iter denotes the average exit step selected by the gate.}
\label{tab:exitgate_q_sweep_140m_hybrid}
\resizebox{\textwidth}{!}{
\begin{tabular}{lccccccccccc}
\toprule
\textbf{Stage}
& \textbf{$q$}
& \textbf{Iter}
& \textbf{PPL}
& \textbf{PIQA}
& \textbf{BoolQ}
& \textbf{WG}
& \textbf{OBQA}
& \textbf{HS}
& \textbf{ARC-C}
& \textbf{ARC-E}
& \textbf{Avg.} \\
\midrule
Stage 1 & 1.00 & 4.00 & 15.07 & 50.71 & 47.58 & 49.80 & 26.40 & 25.36 & 27.90 & 26.26 & 36.29 \\
Stage 1 & 0.80 & 4.00 & 15.07 & 50.71 & 47.61 & 49.33 & 26.40 & 25.28 & 27.99 & 26.26 & 36.23 \\
Stage 1 & 0.60 & 3.55 & 15.05 & 50.76 & 48.01 & 49.25 & 27.00 & 25.61 & 28.24 & 26.56 & 36.49 \\
Stage 1 & 0.40 & 2.81 & 15.07 & 51.74 & 54.77 & 49.96 & 26.80 & 25.29 & 28.16 & 25.34 & 37.44 \\
Stage 1 & 0.20 & 2.35 & 15.66 & 52.39 & 40.55 & 50.20 & 26.20 & 25.29 & 28.50 & 25.84 & 35.57 \\
\midrule
Stage 2 & 1.00 & 4.00 & 15.07 & 50.60 & 47.55 & 50.04 & 26.40 & 25.34 & 27.99 & 26.35 & 36.32 \\
Stage 2 & 0.80 & 3.29 & 15.07 & 50.38 & \textbf{54.80} & 49.25 & 26.40 & 25.39 & 27.73 & \textbf{26.68} & 37.23 \\
Stage 2 & 0.60 & 2.64 & 15.28 & 51.47 & 54.65 & 47.91 & 27.00 & 25.19 & 27.90 & 25.42 & 37.08 \\
Stage 2 & 0.40 & 2.27 & 15.95 & \textbf{52.72} & 38.26 & 50.43 & 26.00 & 25.57 & 28.84 & 25.67 & 35.36 \\
Stage 2 & 0.20 & 1.93 & 17.56 & 52.61 & 37.98 & 49.72 & 24.80 & 25.66 & 28.50 & 26.01 & 35.04 \\
\bottomrule
\end{tabular}
}
\end{table*}

\begin{table*}[t]
\centering
\scriptsize
\caption{Threshold sweep for the 370M Mamba-2 exit-gated model. We report Stage 1 and Stage 2 exit-gate models with $q\in\{0.2,0.4,0.6,0.8,1.0\}$. Iter denotes the average exit step selected by the gate.}
\label{tab:exitgate_q_sweep_370m}
\resizebox{\textwidth}{!}{
\begin{tabular}{lccccccccccc}
\toprule
\textbf{Stage}
& \textbf{$q$}
& \textbf{Iter}
& \textbf{PPL}
& \textbf{PIQA}
& \textbf{BoolQ}
& \textbf{WG}
& \textbf{OBQA}
& \textbf{HS}
& \textbf{ARC-C}
& \textbf{ARC-E}
& \textbf{Avg.} \\
\midrule
Stage 1 & 1.00 & 4.00 & 12.55 & 52.56 & 61.04 & 50.04 & 26.40 & 25.19 & 28.24 & 26.30 & \textbf{38.54} \\
Stage 1 & 0.80 & 4.00 & 12.55 & 52.72 & 61.10 & 50.04 & 26.40 & 25.13 & 27.90 & 26.22 & 38.50 \\
Stage 1 & 0.60 & 3.40 & 12.54 & 52.07 & 59.24 & 49.88 & 24.40 & 25.09 & 26.96 & 26.47 & 37.73 \\
Stage 1 & 0.40 & 2.64 & 12.58 & 50.71 & 55.99 & 49.41 & 24.00 & 25.14 & 26.96 & 26.30 & 36.93 \\
Stage 1 & 0.20 & 1.91 & 13.21 & 51.14 & 57.83 & 49.88 & 24.00 & 24.81 & 28.41 & 26.30 & 37.48 \\
\midrule
Stage 2 & 1.00 & 4.00 & 12.55 & 52.50 & 60.80 & 50.04 & 26.40 & 25.14 & 27.30 & 26.22 & 38.34 \\
Stage 2 & 0.80 & 3.18 & 12.55 & 52.34 & 57.61 & 50.59 & 24.80 & 24.77 & 26.88 & 26.09 & 37.58 \\
Stage 2 & 0.60 & 2.53 & 12.69 & 50.54 & 56.21 & 50.75 & 24.20 & 25.49 & 28.07 & 26.05 & 37.33 \\
Stage 2 & 0.40 & 1.94 & 13.17 & 50.87 & 57.49 & 49.49 & 24.00 & 24.76 & 27.30 & 25.13 & 37.01 \\
Stage 2 & 0.20 & 1.67 & 14.41 & 50.60 & 57.22 & 49.49 & 24.60 & 24.32 & 27.90 & 25.59 & 37.10 \\
\bottomrule
\end{tabular}
}
\end{table*}

\begin{table*}[t]
\centering
\scriptsize
\caption{Threshold sweep for the 370M Mamba-2 Hybrid exit-gated model. We report Stage 1 and Stage 2 exit-gate models with $q\in\{0.2,0.4,0.6,0.8,1.0\}$. Iter denotes the average exit step selected by the gate.}
\label{tab:exitgate_q_sweep_370m_hybrid}
\resizebox{\textwidth}{!}{
\begin{tabular}{lccccccccccc}
\toprule
\textbf{Stage}
& \textbf{$q$}
& \textbf{Iter}
& \textbf{PPL}
& \textbf{PIQA}
& \textbf{BoolQ}
& \textbf{WG}
& \textbf{OBQA}
& \textbf{HS}
& \textbf{ARC-C}
& \textbf{ARC-E}
& \textbf{Avg.} \\
\midrule
Stage 1 & 1.00 & 4.00 & 12.13 & 52.01 & 57.00 & 51.14 & 27.80 & 25.63 & 30.12 & 25.46 & 38.45 \\
Stage 1 & 0.80 & 4.00 & 12.13 & 52.29 & 56.88 & 51.14 & 27.80 & 25.73 & 30.12 & 25.59 & \textbf{38.51} \\
Stage 1 & 0.60 & 3.34 & 12.12 & 51.03 & 39.54 & 50.67 & 27.60 & 26.36 & 30.63 & 25.84 & 35.95 \\
Stage 1 & 0.40 & 2.74 & 12.15 & 51.03 & 39.42 & 49.41 & 28.80 & 25.24 & 31.48 & 25.34 & 35.82 \\
Stage 1 & 0.20 & 2.24 & 12.66 & 51.31 & 42.02 & 50.59 & 29.40 & 25.39 & 31.14 & 25.88 & 36.53 \\
\midrule
Stage 2 & 1.00 & 4.00 & 12.13 & 52.18 & 57.00 & 51.38 & 27.80 & 25.63 & 30.12 & 25.59 & 38.53 \\
Stage 2 & 0.80 & 3.01 & 12.12 & 51.09 & 39.45 & 50.04 & 27.40 & 25.84 & 30.29 & 26.09 & 35.74 \\
Stage 2 & 0.60 & 2.57 & 12.21 & 51.14 & 39.51 & 50.04 & 28.20 & 25.36 & 31.23 & 25.80 & 35.90 \\
Stage 2 & 0.40 & 2.22 & 12.60 & 51.69 & 41.96 & 48.86 & 29.40 & 25.15 & 31.06 & 25.88 & 36.29 \\
Stage 2 & 0.20 & 1.88 & 13.62 & 51.20 & 41.99 & 50.28 & 29.20 & 25.03 & 30.97 & 25.80 & 36.35 \\
\bottomrule
\end{tabular}
}
\end{table*}

\end{document}